\documentclass{article} % For LaTeX2e
\usepackage{iclr2024_conference,times}

\usepackage{amsmath,amsfonts,bm}

\def\eqref#1{equation~\ref{#1}}
\def\1{\bm{1}}

\def\vtheta{{\bm{\theta}}}

\def\vc{{\bm{c}}}

\def\vf{{\bm{f}}}

\def\vx{{\bm{x}}}
\def\vy{{\bm{y}}}
\def\vz{{\bm{z}}}

\DeclareMathAlphabet{\mathsfit}{\encodingdefault}{\sfdefault}{m}{sl}
\SetMathAlphabet{\mathsfit}{bold}{\encodingdefault}{\sfdefault}{bx}{n}

\usepackage{hyperref}
\usepackage{url}

\usepackage{graphicx}
\usepackage{subfigure}
\usepackage{float}
\usepackage{tipa}

\usepackage{bm}

\usepackage{multirow}
\usepackage{amsthm}
\usepackage{wrapfig}
\usepackage{multicol}
\usepackage{algorithm}
\usepackage{algorithmic}
\usepackage{amsmath}
\usepackage{makecell}
\usepackage[textsize=tiny]{todonotes}
\usepackage{booktabs}
\usepackage{amssymb}

\usepackage{enumitem}

\title{Latent Consistency Models: \\ Synthesizing High-Resolution Images \\ with Few-step Inference}
\author{Simian Luo$^*$ \qquad Yiqin Tan\thanks{Equal Contribution\quad $^\dagger$Corresponding Authors} \qquad Longbo Huang$^\dagger$ \qquad Jian Li$^\dagger$ \qquad Hang Zhao$^\dagger$  \\
Institute for Interdisciplinary Information Sciences, Tsinghua University\\
\texttt{\{luosm22, tyq22\}@mails.tsinghua.edu.cn} \\
\texttt{\{longbohuang, lijian83, hangzhao\}@tsinghua.edu.cn}\\
}

\def\@onedot{\ifx\@let@token.\else.\null\fi\xspace}

\makeatother

\iclrfinalcopy % Uncomment for camera-ready version, but NOT for submission.
\begin{document}

\maketitle

\begin{abstract}
Latent Diffusion models (LDMs) have achieved remarkable results in synthesizing high-resolution images. However, the iterative sampling process is computationally intensive and leads to slow generation.
Inspired by Consistency Models~\citep{song2023consistency}, 
we propose Latent Consistency Models (\textbf{LCMs}), enabling swift inference with minimal steps on any pre-trained LDMs, including Stable Diffusion~\citep{rombach2022high}. 
Viewing the guided reverse diffusion process as solving an augmented probability flow ODE (PF-ODE), LCMs are designed to directly predict the solution of such ODE in latent space, mitigating the need for numerous iterations and allowing rapid, high-fidelity sampling. 
Efficiently distilled from pre-trained classifier-free guided diffusion models, a high-quality 768$\times$768 2$\sim$4-step LCM takes only 32 A100 GPU hours for training.
Furthermore, we introduce Latent Consistency Fine-tuning (LCF), a novel method that is tailored for fine-tuning LCMs on customized image datasets. Evaluation on the LAION-5B-Aesthetics dataset demonstrates that LCMs achieve state-of-the-art text-to-image generation performance with few-step inference. Project Page: \url{https://latent-consistency-models.github.io/}

\end{abstract}

\section{Introduction}
\vspace{-5pt}

\begin{figure}[t] 
\begin{centering}
\includegraphics[scale=0.75]{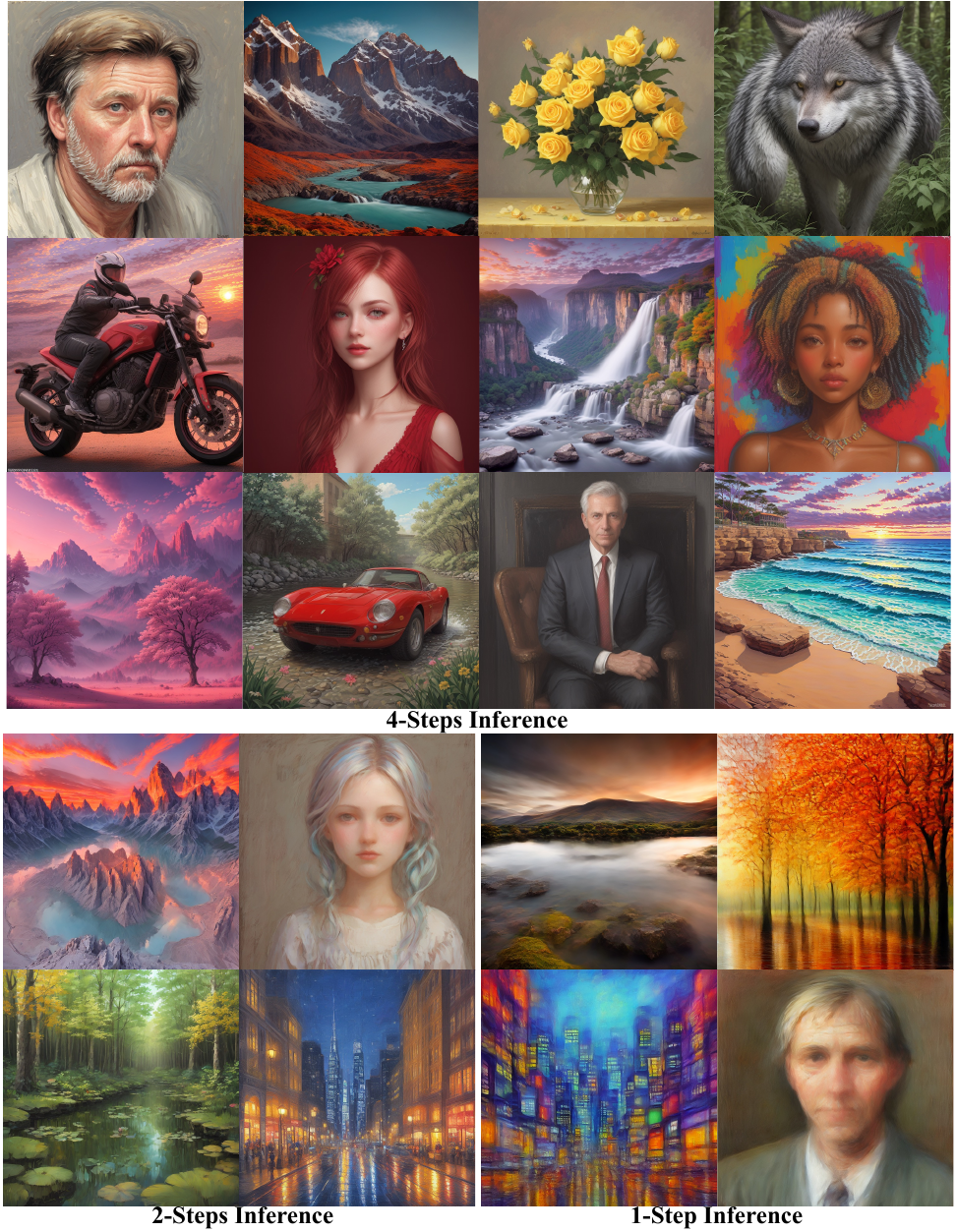}
 \vspace{-0.15in}
\caption{\footnotesize{Images generated by Latent Consistency Models (LCMs) with CFG scale $\omega=8.0$. LCMs can be distilled from any pre-trained Stable Diffusion (SD) in only 4,000 training steps ($\sim$\textbf{32 A100 GPU Hours}) for generating high quality 768$\times$768 resolution images in 2$\sim$4 steps or even one step, significantly accelerating text-to-image generation. We employ LCM to distill the Dreamer-V7 version of SD in just 4,000 training iterations.}\label{fig:teaser}}
\end{centering}
\vspace{-0.26in}
\end{figure}

%\vspace{-1.5cm}
Diffusion models have emerged as powerful generative models that have gained significant attention and achieved remarkable results in various domains \citep{ho2020denoising,song2020denoising,nichol2021improved,ramesh2022hierarchical, song2019generative, song2021maximum}. In particular, latent diffusion models (LDMs) (e.g., Stable Diffusion \citep{rombach2022high}) have demonstrated exceptional performance, especially in high-resolution text-to-image synthesis tasks. LDMs can generate high-quality images conditioned on textual descriptions by utilizing an iterative reverse sampling process that performs gradual denoising of samples. However, diffusion models suffer from a notable drawback: the iterative reverse sampling process leads to slow generation speed, limiting their real-time applicability. To overcome this drawback, researchers have proposed several methods to improve the sampling speed, which involves accelerating the denoising process by enhancing ODE solvers \citep{ho2020denoising,lu2022dpm,lu2022dpm++}, which can generate images within 10$\sim$20 sampling steps. Another approach is to distill a pre-trained diffusion model into models that enable few-step inference \cite{salimans2022progressive,meng2023distillation}. In particular, \cite{meng2023distillation} proposed a two-stage distillation approach to improving the sampling efficiency of classifier-free guided models.
Recently, \cite{song2023consistency} proposed consistency models as a promising alternative aimed at speeding up the generation process. By learning consistency mappings that maintain point consistency on ODE-trajectory, these models allow for single-step generation, eliminating the need for computation-intensive iterations. However, \cite{song2023consistency} is constrained to pixel space image generation tasks, making it unsuitable for synthesizing high-resolution images. Moreover, the applications to the conditional diffusion model and the incorporation of classifier-free guidance have not been explored, rendering their methods unsuitable for text-to-image generation synthesis. 

% Further, the applications to conditional diffusion model and the incorporation of classifier-free guidance have not been explored, rendering their methods inapplicable for  text-to-image generation synthesis.

% In this paper, we introduce ``Latent Consistency Models" (LCM), a new approach for fast generation on text-to-image tasks. As in latent diffusion models, we introduce a latent space which captures meaningful representations of the data, thus providing a compact and efficient representation of the generative process. 
% \jian{I feel the latent space is not exactly the novalty of our work.
% how about the following: Mirroring latent diffusion models, we employ consistency models in the latent space of a pre-trained autoencoder, which is a lower-dimensional (thereby more spatially efficient) representational space capturing the complex data distribution.}
% We propose an one-stage guided distillation method which converts a pre-trained guided diffusion model to a latent consistency model, which can be done efficiently by solving an Augmented PF-ODE and adopting \textsc{Skipping-Step} technique. We further explore a new fine-tuning paradigm, named Consistency Fine-tuning, which allow fine-tuning a pre-trained latent consistency model to support one-step or few-step ($2\sim4$) generation on customized image datasets.

In this paper, we introduce \textbf{Latent Consistency Models} (LCMs) for fast, high-resolution image generation. Mirroring LDMs, we employ consistency models in the image latent space of a pre-trained auto-encoder from Stable Diffusion \citep{rombach2022high}.
% , which is a lower-dimensional (thereby more spatially efficient) representational space capturing the complex data distribution.}
We propose a one-stage guided distillation method to efficiently convert a pre-trained guided diffusion model into a latent consistency model by solving an augmented PF-ODE. Additionally, we propose Latent Consistency Fine-tuning, which allows fine-tuning a pre-trained LCM to support few-step inference on customized image datasets.
Our main contributions are summarized as follows:

\vspace{-2pt}\begin{itemize}[leftmargin=*, itemsep=1pt]
    \item We propose Latent Consistency Models (LCMs) for fast, high-resolution image generation.
    LCMs employ consistency models in the image latent space, enabling fast few-step or even one-step high-fidelity sampling on pre-trained latent diffusion models (e.g., Stable Diffusion (SD)). 
    \item  We provide a simple and efficient one-stage {\em guided consistency distillation} method to distill SD for few-step (2$\sim$4) or even 1-step sampling. 
    We propose the \textsc{Skipping-Step} technique to further accelerate the convergence.
    For 2- and 4-step inference, our method costs only 32 A100 GPU hours for training and 
    achieves state-of-the-art performance on the LAION-5B-Aesthetics dataset.
    \item We introduce a new fine-tuning method for LCMs, named Latent Consistency Fine-tuning, enabling efficient adaptation of a pre-trained LCM to customized datasets while preserving the ability of fast inference.
\end{itemize}

% (Jian: we may need a summary of contributions in the introduction)

\vspace{-10pt}
\section{Related Work}
\vspace{-5pt}
% In this section, we discuss previous works on generative models, including diffusion model and consistency model. 

\noindent \textbf{Diffusion Models} have achieved great success in image generation  \citep{ho2020denoising,song2020denoising,nichol2021improved,ramesh2022hierarchical,rombach2022high, song2019generative}.
They are trained to denoise the noise-corrupted data to estimate the {\em score} of data distribution.
During inference, samples are drawn by running the reverse diffusion process to gradually denoise the data point. Compared to VAEs \citep{kingma2013auto, sohn2015learning} and GANs \citep{goodfellow2020generative}, diffusion models enjoy the benefit of training stability and better likelihood estimation. 

\noindent \textbf{Accelerating DMs.} 
However, diffusion models are bottlenecked by their slow generation speed. Various approaches have been proposed, including training-free methods such as ODE solvers \citep{song2020denoising, lu2022dpm, lu2022dpm++}, adaptive step size solvers \citep{jolicoeur2021gotta}, predictor-corrector methods \citep{song2020score}. Training-based approaches include optimized discretization \citep{watson2021learning}, truncated diffusion \citep{lyu2022accelerating,zheng2022truncated}, neural operator \citep{zheng2023fast} and distillation \citep{salimans2022progressive,meng2023distillation}. More recently, new generative models for faster sampling have also been proposed \citep{liu2022flow,liu2023instaflow}.

\noindent \textbf{Latent Diffusion Models} (LDMs) \citep{rombach2022high} excel in synthesizing high-resolution text-to-images. For example, Stable Diffusion (SD) performs forward and reverse diffusion processes in the data latent space, resulting in more efficient computation. 
% The cross-attention in LDMs that encode text empowers the models to synthesize image that aligns well with the text. 
% Thanks to LDMs, SD can run inference on the GPUs of personal laptops.

\noindent \textbf{Consistency Models} (CMs) \citep{song2023consistency} have shown great potential as a new type of generative model for faster sampling while preserving generation quality. 
%Specifically, while diffusion models excel in generating high-quality samples, the iterative sampling process leads to slow generation. In contrast, 
CMs adopt consistency mapping to directly map any point in ODE trajectory to its origin, enabling fast one-step generation. 
CMs can be trained by distilling pre-trained diffusion models or as standalone generative models. Details of CMs are elaborated in the following section.
% CMs outperform other one-step, non-adversarial generative models on standard benchmarks, making them a substantial contribution to the field of generative modeling. 
% \vspace{-0.3in}

\vspace{-10pt}
\section{Preliminaries}
\vspace{-5pt}
%In this section, we provide an overview of the diffusion models and consistency models.
In this section, we briefly review diffusion and consistency models and define relevant notations.

\noindent
{\bf Diffusion Models:}
Diffusion models, or score-based generative models \cite{ho2020denoising, song2020denoising} is a family of generative models that progressively inject Gaussian noises into the data, and then generate samples from noise via a reverse denoising process.
In particular, diffusion models define a forward process transitioning the origin data distribution $p_{data}(x)$ to marginal distribution $q_t(\boldsymbol{x}_t)$, via transition kernel: \begin{small}$ q_{0 t}\left(\boldsymbol{x}_t \mid \boldsymbol{x}_0\right)=\mathcal{N}\left(\boldsymbol{x}_t \mid \alpha(t) \boldsymbol{x}_0, \sigma^2(t) \boldsymbol{I}\right)$\end{small}, where $\alpha(t), \sigma(t)$ specify the noise schedule. In continuous time perspective, the forward process can be described by a stochastic differential equation (SDE) \cite{song2020score, lu2022dpm, karras2022elucidating} for $t\in[0, T]$:
%\begin{small}
%\begin{equation}
$
    \mathrm{d} \boldsymbol{x}_t=f(t) \boldsymbol{x}_t \mathrm{~d} t+g(t) \mathrm{d} \boldsymbol{w}_t, \,\, \boldsymbol{x}_0 \sim p_{data}\left(\boldsymbol{x}_0\right),
$
%\end{equation}
%\end{small}
where $\boldsymbol{w}_t$ is the standard Brownian motion, and
\begin{small}
\begin{equation}
    f(t)=\frac{\mathrm{d} \log \alpha(t)}{\mathrm{~d} t}, \quad g^2(t)=\frac{\mathrm{d} \sigma^2(t)}{\mathrm{~d} t}-2 \frac{\mathrm{d} \log \alpha(t)}{\mathrm{~d} t} \sigma^2(t).
\end{equation}
\end{small}
% \jian{$\alpha_t$ and $\alpha(t)$. better be consistent.}
%Leveraging the classic result of Anderson (1982), \cite{song2020score} show that reversing the forward process is also a diffusion process, specified by
%the following reverse-time SDE for data sampling from $T$ to $0$, starting with $q_T(\boldsymbol{x}_T)$: 
%\begin{small}
%\begin{equation}
%    \mathrm{d} \boldsymbol{x}_t=\left[f(t) \boldsymbol{x}_t-g^2(t) \nabla_{\boldsymbol{x}} \log %q_t\left(\boldsymbol{x}_t\right)\right] \mathrm{d} t+g(t) \mathrm{d} \overline{\boldsymbol{w}}_t, %\quad \boldsymbol{x}_T \sim q_T\left(\boldsymbol{x}_T\right) \label{eq:backward_sde},
%\end{equation}
%\end{small}
%where $\overline{\boldsymbol{w}}_t$ is a standard reverse-time Brownian motion.
%It is known that for the above SDE (Eq.\ref{eq:backward_sde}), 
By considering the reverse time SDE (see Appendix~\ref{app:preliminary} for more details), one can show that the marginal distribution $q_t(\boldsymbol{x})$ satisfies the following ordinary differential equation, called the \textit{Probability Flow ODE} (PF-ODE) \citep{song2020score, lu2022dpm}:
\begin{small}
\begin{equation}
    \frac{\mathrm{d} \boldsymbol{x}_t}{\mathrm{~d} t}=f(t) \boldsymbol{x}_t-\frac{1}{2} g^2(t) \nabla_{\boldsymbol{x}} \log q_t\left(\boldsymbol{x}_t\right), \,\, \boldsymbol{x}_T \sim q_T\left(\boldsymbol{x}_T\right). \label{eq:pf_origin_ode}
\end{equation}
\end{small}
In diffusion models, we train the noise prediction model $\boldsymbol{\epsilon}_\theta(\boldsymbol{x}_t, t)$ to fit 
$-\nabla \log q_t(\boldsymbol{x}_t)$ (called the {\em score function}).
Approximating the score function by the noise prediction model in \ref{eq:pf_origin_ode},
one can obtain the following {\em empirical PF-ODE} for sampling:
\begin{small}
\begin{equation}
    \frac{\mathrm{d} \boldsymbol{x}_t}{\mathrm{~d} t}= f(t) \boldsymbol{x}_t+\frac{g^2(t)}{2 \sigma_t} \boldsymbol{\epsilon}_\theta\left(\boldsymbol{x}_t, t\right), \quad \boldsymbol{x}_T \sim \mathcal{N}\left(\mathbf{0}, \tilde{\sigma}^2 \boldsymbol{I}\right).
    \label{eq:pf_empirical_ode}
\end{equation}
\end{small}
For class-conditioned diffusion models, Classifier-Free Guidance (CFG) \citep{ho2022classifier} is an effective technique to significantly improve the quality of generated samples 
and has been widely used in several large-scale diffusion models including GLIDE \cite{Alex2021glide}, Stable Diffusion \citep{rombach2022high}, DALL·E 2 \citep{ramesh2022hierarchical} and Imagen \citep{saharia2022photorealistic}. Given a CFG scale $\omega$, the original noise prediction is replaced by a linear combination of conditional and unconditional noise prediction, i.e., $\bm{\tilde{\epsilon}_\theta}(\vz_t,\omega,\vc,t)=(1+\omega)\bm{\epsilon_\theta}(\vz_t,\vc,t)-\omega\bm{\epsilon_\theta}(\vz,\varnothing,t).$

\vspace{0.1cm}

\noindent
{\bf Consistency Models:}
The Consistency Model (CM) \citep{song2023consistency} is a new family of generative models that enables one-step or few-step generation. The core idea of the CM is to learn the function that maps any points on a trajectory of the PF-ODE to that trajectory's origin (i.e., the solution of the PF-ODE). More formally, the consistency function is defined as
%\begin{small}
%\begin{equation}
$
    \vf:(\vx_t,t)\longmapsto \vx_\epsilon,
$
%\end{equation}
%\end{small}
where $\epsilon$ is a fixed small positive number. 
%In other words, the output of the $\vf_\vtheta(\vx, t)$ should be the solution of the PF-ODE: $\boldsymbol{x}_\epsilon= \text{Solve (PF-ODE)} =\vf_\vtheta(\vx, t)$.
% \jian{shouldn't be $\vx_\epsilon$? be consistent.}
One important observation is that the consistency function should satisfy the 
{\em self-consistency property}:
\begin{small}
\begin{equation}
    \vf(\vx_t,t)=\vf(\vx_{t'},t'),\forall t,t'\in[\epsilon,T]. \label{eq:self_consistency}
\end{equation}
\end{small}
The key idea in \citep{song2023consistency} for learning a consistency model $\vf_\vtheta$ is to learn a consistency function from data by effectively enforcing the self-consistency property in Eq.~\ref{eq:self_consistency}. 
To ensure that $\vf_\vtheta(\vx,\epsilon)=\vx$, 
the consistency model $\vf_\vtheta$ is parameterized as:
\begin{small}
\begin{equation}
    \vf_\vtheta(\vx,t)=c_{\text{skip}}(t)\vx+c_{\text{out}}(t)\boldsymbol{F}_\vtheta(\vx,t),
\end{equation}
\end{small}
where \begin{small}$c_{\text{skip}}(t)$\end{small} and \begin{small}$c_{\text{out}}(t)$\end{small} are differentiable functions with \begin{small}$c_{\text{skip}}(\epsilon)=1$\end{small} and \begin{small}$c_{\text{out}}(\epsilon)=0$\end{small},
and $\boldsymbol{F}_\vtheta(\vx,t)$ is a deep neural network. A CM can be either distilled from a pre-trained diffusion model or trained from scratch. The former is known as {\em Consistency Distillation}. 
To enforce the self-consistency property,
we maintain a target model $\vtheta^-$, updated with exponential moving average (EMA) of the parameter $\vtheta$ we intend to learn, i.e., \begin{small}$\vtheta^-\leftarrow\mu\vtheta^-+(1-\mu)\vtheta$\end{small},
% \begin{small}
% \begin{equation}
%     \vtheta^-\leftarrow\mu\vtheta^-+(1-\mu)\vtheta. \label{eq:ema}
% \end{equation}
% \end{small}
and define the consistency loss as follows:
\begin{small}
\begin{equation}
    \mathcal{L}(\vtheta, \vtheta^-;\Phi)=\mathbb{E}_{\boldsymbol{x},t}
    \left[d\left(\vf_\vtheta(\vx_{t_{n+1}},t_{n+1}),\vf_{\vtheta^{-}}(\hat{\vx}^\phi_{t_n},t_n)\right)\right],\label{eq:consistency_loss}
\end{equation}
\end{small}
where $d(\cdot,\cdot)$ is a chosen metric function for measuring the distance between two samples, e.g., the squared $\ell_2$ distance \begin{small}$d(\vx,\vy)=||\vx-\vy||_2^2$\end{small}. 
$\hat{\vx}_{t_n}^\phi$ is a one-step estimation of $\boldsymbol{x}_{t_n}$ from $\boldsymbol{x}_{t_{n+1}}$ as:
\begin{equation}
    \hat{\vx}_{t_n}^\phi\leftarrow \vx_{t_{n+1}}+(t_n-t_{n+1})\Phi(\vx_{t_{n+1}},t_{n+1};\phi).
\end{equation}
where $\Phi$ denotes the one-step ODE solver applied to PF-ODE in Eq.~\ref{eq:simple_pf_ode}. \citep{song2023consistency} used Euler \citep{song2020score} or Heun solver \citep{karras2022elucidating} as the numerical ODE solver.
More details and the pseudo-code for consistency distillation (Algorithm~\ref{alg:CD_raw}) 
are provided in Appendix~\ref{app:preliminary}.

 % \jian{need to specify the formula for $t_i$.}

% The parameter $\vtheta$ is learnt using stochastic gradient descent until convergence. The pseudo-code is presented in Algorithm~\ref{alg:CD_raw} in Appendix~\ref{xxx}.

%\begin{multicols}{2}

\vspace{-10pt}
\section{Latent Consistency Models}
\vspace{-5pt}

Consistency Models (CMs)~\citep{song2023consistency} only focused on image generation tasks on ImageNet 64$\times$64 \citep{deng2009imagenet} and LSUN 256$\times$256 \citep{yu2015lsun}. The potential of CMs to generate higher-resolution text-to-image tasks remains unexplored. In this paper, we introduce \textbf{Latent Consistency Models} (LCMs) in Sec~\ref{sec:lcd} to tackle these more challenging tasks, unleashing the potential of CMs. Similar to LDMs, our LCMs adopt a consistency model in the image latent space. We choose the powerful Stable Diffusion (SD) as the underlying diffusion model to distill from. 
%a widely used LDM known for its exceptional ability to generate high-resolution images.
We aim to achieve few-step (2$\sim$4) and even one-step inference on SD without compromising image quality. The classifier-free guidance (CFG) \citep{ho2022classifier} is an effective technique to further improve sample quality and is widely used in SD. However, its application in CMs remains unexplored. We propose a simple one-stage guided distillation method in Sec~\ref{sec:aug_pf_ode} that solves an {\em augmented PF-ODE}, integrating CFG into LCM effectively. We propose \textsc{Skipping-Step} technique to accelerate the convergence of LCMs in Sec.~\ref{sec:skipping_step}. Finally, we propose Latent Consistency Fine-tuning to finetune a pre-trained LCM for few-step inference on a customized dataset in Sec~\ref{sec:consitency_finetune}.

\vspace{-10pt}
\subsection{Consistency Distillation in the Latent Space \label{sec:lcd}}
\vspace{-5pt}
%Exploiting the latent space of images to improve image generation quality, while simultaneously reducing computational overhead, has been demonstrably successful in prior research. 
%Models such as the Latent Diffusion Model (LDM), exemplified by Stable Diffusion \cite{rombach2022high}, first train an autoencoder ($\mathcal{E}, \mathcal{D}$) that compress the origin high-dimensional image data $x$ into a low-dimensional latent $z = \mathcal{E}(x)$ , which is then decoded to reconstruct the image as $\hat{x} = \mathcal{D}(z)$. By training diffusion models on image latent space, computational cost are substantially reduced compared to direct training on high-resolution images. Additionally, this approach accelerates inference processes, enabling users to proficiently generate high-resolution images on personal laptop GPUs.

% Leveraging the latent space of images to enhance image generation quality, while  reducing the computational overhead, has proven to remarkably successful in large-scale diffusion models. 
Utilizing image latent space in large-scale diffusion models like Stable Diffusion (SD) \citep{rombach2022high} has effectively enhanced image generation quality and reduced computational load. In SD, an autoencoder ($\mathcal{E}, \mathcal{D}$) is first trained to compress high-dim image data into low-dim latent vector $z=\mathcal{E}(x)$, which is then decoded to reconstruct the image as $\hat{x}=\mathcal{D}(z)$. Training diffusion models in the latent space greatly reduces the computation costs compared to pixel-based models and speeds up the inference process; LDMs make it possible to generate high-resolution images on laptop GPUs.
% Such latent diffusion models (e.g., Stable Diffusion \cite{rombach2022high}) initially train an autoencoder (($\mathcal{E}, \mathcal{D}$)
% to compress the original high-dimensional image data 
% into a low-dimensional latent vector $z = \mathcal{E}(x)$, 
% which is then decoded to reconstruct the image as $\hat{x} = \mathcal{D}(z)$. 
% By training diffusion models on the latent space, computational costs are substantially reduced compared to direct training on high-resolution images. Moreover, this strategy significantly accelerates the inference process, enabling generating of high-resolution images even on personal laptop GPUs.
For LCMs, we leverage the advantage of the latent space for consistency distillation, contrasting with the pixel space used in CMs \citep{song2023consistency}. This approach, termed \textbf{Latent Consistency Distillation (LCD)} is applied to pre-trained SD, allowing the synthesis of high-resolution (e.g., 768$\times$768) images in 1$\sim$4 steps.
% In our implementation, we apply LCD to the pre-trained Stable Diffusion models \citep{rombach2022high}, enabling us to synthesize high resolution (e.g , $512\times512$ , $768\times768$) images within a few-step ($2\sim4$) and even $1$ step. 
We focus on conditional generation. Recall that the PF-ODE of the reverse diffusion process \citep{song2020score, lu2022dpm} is
\begin{small}
\begin{equation}
    \frac{\mathrm{d} \boldsymbol{\vz}_t}{\mathrm{~d} t}= f(t) \boldsymbol{\vz}_t+\frac{g^2(t)}{2 \sigma_t} \boldsymbol{\epsilon}_\theta\left(\boldsymbol{\vz}_t, \boldsymbol{c}, t\right), \quad \boldsymbol{\vz}_T \sim \mathcal{N}\left(\mathbf{0}, \tilde{\sigma}^2 \boldsymbol{I}\right),
    \label{eq:pf_ode_vanilla}
\end{equation}
\end{small}
where $\boldsymbol{\vz}_t$ are image latents, $\boldsymbol{\epsilon}_\theta\left(\boldsymbol{\vz}_t, \boldsymbol{c}, t\right)$ is the noise prediction model, and $\boldsymbol{c}$ is the given condition (e.g text). Samples can be drawn by solving the PF-ODE from $T$ to $0$. To perform \textbf{LCD}, we introduce the consistency function 
$
\boldsymbol{f_{\theta}}: (\boldsymbol{z_t}, \boldsymbol{c}, t) \mapsto \boldsymbol{z_0}
$
to directly predict the solution of {\em{PF-ODE}} (Eq.~\ref{eq:pf_ode_vanilla}) for $t=0$. 
We parameterize $\boldsymbol{f_{\theta}}$ by the noise prediction model $\hat{\boldsymbol{\epsilon}}_\theta$, as follows:
\begin{footnotesize}
\begin{equation}
    \vf_\vtheta(\vz, \vc, t)=c_{\text{skip}}(t)\vz+c_{\text{out}}(t)\left(\frac{\vz - \sigma_t \hat{\boldsymbol{\epsilon}}_\theta(\vz, \vc,t)}{\alpha_t}\right),\quad (\boldsymbol{\epsilon}\text{-Prediction})
    \label{eq:parameterization}
% \vspace{-0.04in}
\end{equation}
\end{footnotesize}
where \begin{small}$c_{\text{skip}}(0)=1, c_{\text{out}}(0)=0$\end{small}
and $\hat{\boldsymbol{\epsilon}}_\theta(\vz, \vc,t)$ is a noise prediction model that initializes with the same parameters as the teacher diffusion model. Notably, $\boldsymbol{f_{\theta}}$ can be parameterized in various ways, depending on the teacher diffusion model parameterizations of predictions (e.g., $\boldsymbol{x}$, $\boldsymbol{\epsilon}$ \citep{ho2020denoising}, $\boldsymbol{v}$ \citep{salimans2022progressive}). We discuss other possible parameterizations in Appendix~\ref{appendix:parameterization}.

% where \begin{small}$c_{\text{skip}}(0)=1, c_{\text{out}}(0)=0$\end{small}
% and $\hat{\boldsymbol{\epsilon}}_\theta(\vz, \vc,t)$ is a noise prediction model that initializes with the same parameters as the teacher diffusion model. Notably, $\boldsymbol{f_{\theta}}$ can be parameterized differently, depending on the teacher model's parameterizations. We discuss other possible parameterizations in \lsm{Appendix}.
We assume that an efficient ODE solver $\Psi(\vz_{t},t, s, \vc)$ is available for approximating the integration
of the right-hand side of Eq~\eqref{eq:pf_ode_vanilla}
from time $t$ to $s$.
In practice, we can use DDIM \citep{song2020denoising}, DPM-Solver \citep{lu2022dpm} or DPM-Solver++ \citep{lu2022dpm++} as $\Psi(\cdot,\cdot,\cdot,\cdot)$.
Note that we only use these solvers in training/distillation, not in inference.
% \footnote{These solvers can be of course directly used for inference, but they typically require many steps.}
We will discuss these solvers further when we introduce the \textsc{skipping-step} technique in Sec.~\ref{sec:skipping_step}. LCM aims to predict the solution of the PF-ODE by minimizing the consistency distillation loss \citep{song2023consistency}: 
\begin{small}
% \vspace{-0.08in}
\begin{equation}
\mathcal{L_{CD}}\left(\boldsymbol{\theta},\boldsymbol{\theta^-};\Psi\right) = \mathbb{E}_{\vz,\boldsymbol{c},n}\left[d\left(\boldsymbol{f_\theta}(\boldsymbol{\vz}_{t_{n+1}}, \boldsymbol{c}, t_{n+1})\ , \boldsymbol{f_{\theta^-}}(\boldsymbol{\hat{\vz}}^{\Psi}_{t_{n}},  \boldsymbol{c}, t_{n})\right)\right].
\label{eq:lcm_consistency_loss_vanilla}
\vspace{-0.05in}
\end{equation}
\end{small}
Here, $\boldsymbol{\hat{\vz}}^{\Psi}_{t_{n}}$ is an estimation of the evolution of the {\em{PF-ODE}} from $t_{n+1}\rightarrow t_{n}$ using ODE solver $\Psi$:
\begin{footnotesize}
\begin{equation}
\begin{aligned}
    \hat{\vz}^{\Psi}_{t_n} - \boldsymbol{\vz}_{t_{n+1}} =  \int_{t_{n+1}}^{t_n}\left(f(t) \boldsymbol{\vz}_t+\frac{g^2(t)}{2 \sigma_t} \boldsymbol{\epsilon}_{\theta}\left(\boldsymbol{\vz}_t, \boldsymbol{c},  t\right)\right) \mathrm{d}t 
    \approx  \Psi(\vz_{t_{n+1}},t_{n+1}, t_n, \vc),
    \label{eq:aug_pf_ode_estimate_vanilla}
\end{aligned}
\vspace{-0.03in}
\end{equation}
\end{footnotesize}
where the solver $\Psi(\cdot,\cdot,\cdot,\cdot)$ is used to approximate the integration from $t_{n+1}\rightarrow t_{n}$.
 
\vspace{-10pt}
\subsection{One-Stage Guided Distillation by solving augmented PF-ODE \label{sec:aug_pf_ode}}
\vspace{-5pt}
Classifier-free guidance (CFG) \citep{ho2022classifier} is crucial for synthesizing high-quality text-aligned images in SD, typically needing a CFG scale $\omega$ over $6$. Thus, integrating CFG into a distillation method becomes indispensable. Previous method Guided-Distill \citep{meng2023distillation} introduces a two-stage distillation to support few-step sampling from a guided diffusion model. However, it is computationally intensive (e.g. at least \textbf{45} A100 GPUs \textbf{Days} for 2-step inference, estimated in \citep{liu2023instaflow}). An LCM demands merely \textbf{32} A100 GPUs \textbf{Hours} training for 2-step inference, as depicted in Figure~\ref{fig:teaser}. Furthermore, the two-stage guided distillation might result in accumulated error, leading to suboptimal performance. In contrast, LCMs adopt efficient one-stage guided distillation by solving an augmented PF-ODE. 
% Classifier-free guidance (CFG) \citep{ho2022classifier} has proven effective in synthesizing realistic images that closely align with text prompts. To generate high-quality images through Stable Diffusion \citep{rombach2022high}, a CFG scale $\omega$ larger than $6$ is typically needed. Consequently, integrating classifier-free guidance into a distillation method becomes indispensable. \cite{meng2023distillation} proposed a two-stage method, called Guided-Distill, to support few-step sampling from a guided diffusion model. 
% However, their approach requires long training hours (e.g., \jian{how long? an example here}) and leads to accumulation error between the two stages. 
% In contrast, we propose an efficient one-stage guided distillation method by solving an augmented probability flow ODE (Augmented PF-ODE). 
%First, we recall that the PF-ODE of the reverse diffusion process is \citep{song2020score, lu2022dpm}
%\begin{equation}
%    \frac{\mathrm{d} \boldsymbol{\vz}_t}{\mathrm{~d} t}= f(t) \boldsymbol{\vz}_t+\frac{g^2(t)}{2 \sigma_t} \boldsymbol{\epsilon}_\theta\left(\boldsymbol{\vz}_t, \boldsymbol{c}, t\right), \quad \boldsymbol{\vz}_T \sim \mathcal{N}\left(\mathbf{0}, \tilde{\sigma}^2 \boldsymbol{I}\right),
%    \label{eq:pf_ode}
%\end{equation}
%where $\boldsymbol{\epsilon}_\theta\left(\boldsymbol{\vz}_t, \boldsymbol{c}, t\right)$ is the noise prediction model, and $\boldsymbol{c}$ is the given condition. Samples can be drawn by solving the PF-ODE from $T$ to $0$. 
Recall the CFG used in reverse diffusion process:
\begin{footnotesize}
\begin{equation}
    \boldsymbol{\tilde{\epsilon}}_{\theta}\left(\boldsymbol{\vz}_t, \omega, \boldsymbol{c}, t\right):= (1 + \omega) \boldsymbol{\epsilon}_\theta\left(\boldsymbol{\vz}_t, \boldsymbol{c}, t\right) - \omega \boldsymbol{\epsilon}_\theta\left(\boldsymbol{\vz}_t, \varnothing, t\right),
    % \vspace{-0.05in}
\end{equation}
\end{footnotesize}
where the original noise prediction is replaced by the linear combination of conditional and unconditional noise and  $\omega$ is called the {\em guidance scale}. To sample from the guided reverse process, we need to solve
the following {\em augmented PF-ODE}: (i.e., augmented with the terms related to $\omega$)
\begin{small}
\begin{equation}
    \frac{\mathrm{d} \boldsymbol{\vz}_t}{\mathrm{~d} t}= f(t) \boldsymbol{\vz}_t+\frac{g^2(t)}{2 \sigma_t} \boldsymbol{\tilde{\epsilon}}_{\theta}\left(\boldsymbol{\vz}_t, \omega, \boldsymbol{c},  t\right), \quad \boldsymbol{\vz}_T \sim \mathcal{N}\left(\mathbf{0}, \tilde{\sigma}^2 \boldsymbol{I}\right).
    \label{eq:aug_pf_ode}
\end{equation}
\end{small}
To efficiently perform one-stage guided distillation, we introduce
an {\em augmented consistency function} 
$\boldsymbol{f_{\theta}}: (\boldsymbol{z_t}, \omega, \boldsymbol{c}, t) \mapsto \boldsymbol{z_0}$
to directly predict the solution of {\em{augmented PF-ODE}} (Eq.~\ref{eq:aug_pf_ode}) for $t=0$. We parameterize the $\boldsymbol{f_{\theta}}$
in the same way as in Eq. \ref{eq:parameterization}, except that
$\hat{\boldsymbol{\epsilon}}_\theta(\vz,\vc,t)$ is replaced by 
$\hat{\boldsymbol{\epsilon}}_\theta(\vz,\omega, \vc,t)$, which is a noise prediction model initializing with the same parameters as the teacher diffusion model, but also contains additional trainable parameters for conditioning on $\omega$. The consistency loss is 
the same as Eq.~\ref{eq:lcm_consistency_loss_vanilla}
except that we use augmented consistency function
$\boldsymbol{f_{\theta}}(\boldsymbol{z_t}, \omega, \boldsymbol{c}, t)$. 
\begin{small}
\begin{equation}
\mathcal{L_{CD}}\left(\boldsymbol{\theta},\boldsymbol{\theta^-};\Psi\right) = \mathbb{E}_{\boldsymbol{\vz,c},\omega,n}\left[d\left(\boldsymbol{f_\theta}(\boldsymbol{\vz}_{t_{n+1}}, \omega, \boldsymbol{c}, t_{n+1})\ , \boldsymbol{f_{\theta^-}}(\boldsymbol{\hat{\vz}}^{\Psi,\omega}_{t_{n}}, \omega, \boldsymbol{c}, t_{n})\right)\right]
\label{eq:lcm_consistency_loss_after}
\end{equation}
\end{small}
In Eq~\ref{eq:lcm_consistency_loss_after}, $\omega$ and $n$ are uniformly sampled from
interval $[\omega_{\text{min}},\omega_{\text{max}}]$ and $\{1,\ldots, N-1\}$ respectively.
$\boldsymbol{\hat{\vz}}^{\Psi,\omega}_{t_{n}}$ is estimated using the new noise model 
$\boldsymbol{\tilde{\epsilon}}_{\theta}\left(\boldsymbol{\vz}_t, \omega, \boldsymbol{c},  t\right)$, as follows:
\begin{footnotesize}
\begin{equation}
\begin{aligned}
    \hat{\vz}^{\Psi, \omega}_{t_n} - \boldsymbol{\vz}_{t_{n+1}} = & \ \int_{t_{n+1}}^{t_n}\left(f(t) \boldsymbol{\vz}_t+\frac{g^2(t)}{2 \sigma_t} \boldsymbol{\tilde{\epsilon}}_{\theta}\left(\boldsymbol{\vz}_t, \omega, \boldsymbol{c},  t\right)\right) \mathrm{d}t \\
    =& \ (1+\omega)\int_{t_{n+1}}^{t_n}\left(f(t) \boldsymbol{\vz}_t+\frac{g^2(t)}{2 \sigma_t} \boldsymbol{\epsilon}_{\theta}\left(\boldsymbol{\vz}_t, \boldsymbol{c},  t\right)\right) \mathrm{d}t  - \omega\int_{t_{n+1}}^{t_n}\left(f(t) \boldsymbol{\vz}_t+\frac{g^2(t)}{2 \sigma_t} \boldsymbol{\epsilon}_{\theta}\left(\boldsymbol{\vz}_t, \varnothing,  t\right)\right) \mathrm{d}t\\
    \approx & \  (1+\omega)\Psi(\vz_{t_{n+1}},t_{n+1}, t_n, \vc)- \omega\Psi(\vz_{t_{n+1}},t_{n+1}, t_n, \varnothing).
    \label{eq:aug_pf_ode_estimate}
\end{aligned}
\end{equation}
\end{footnotesize}
%where we use a {\em{PF-ODE}} Solver $\Psi(\cdot,\cdot,\cdot,\cdot)$ to approximate the integration of {\em{augmented PF-ODE}} from $t_{n+1}\rightarrow t_{n}$. 
Again, we can use DDIM \citep{song2020denoising}, DPM-Solver \citep{lu2022dpm} or DPM-Solver++ \citep{lu2022dpm++} as the PF-ODE solver $\Psi(\cdot,\cdot,\cdot,\cdot)$.
%as discussed in Sec~\ref{sec:skipping_step}.
% , where $\boldsymbol{\hat{\vz}}^{\Psi,\omega}_{t_{n}}$ is a numerical {\em{Augmented PF-ODE}} solver $\Psi$ update from $t_{n+1} \rightarrow t_{n}$.
% \begin{equation}
%     \hat{\vz}^{\Psi, w}_{t_n}\longleftarrow (1+w)\Psi(\vz_{t_{n+1}},t_{n+1}, t_n, \vc)-w\Psi(\vz_{t_{n+1}},t_{n+1}, t_n, \varnothing)
% \end{equation}
% \jian{check the above formula. missing $\hat{\vz}_{t_{n+1}}$?
% also say a few words about the solver $\Psi$. How or where can one get it.}

\vspace{-10pt}
\subsection{Accelerating Distillation with Skipping Time Steps \label{sec:skipping_step}}
\vspace{-5pt}
Discrete diffusion models \citep{ho2020denoising, song2019generative} typically train noise prediction models with a long time-step schedule $\{t_i\}_i$ (also called discretization schedule or time schedule) to achieve high quality generation results. For instance, Stable Diffusion (SD) has a time schedule of length 1,000. However, directly applying Latent Consistency Distillation (\textbf{LCD}) to SD with such an extended schedule can be problematic. The model needs to sample across all 1,000 time steps, and the consistency loss attempts to aligns the prediction of LCM model \begin{small}$\boldsymbol{f_\theta}(\boldsymbol{\vz}_{t_{n+1}}, \boldsymbol{c}, t_{n+1})$\end{small} with
the prediction \begin{small}$\boldsymbol{f_\theta}(\boldsymbol{\vz}_{t_n}, \boldsymbol{c}, t_{n})$\end{small} at the subsequent step along the same trajectory. Since $t_n-t_{n+1}$ is tiny, \begin{small}$\boldsymbol{\vz}_{t_{n}}$\end{small} and \begin{small}$\boldsymbol{\vz}_{t_{n+1}}$\end{small} (and thus \begin{small}$\boldsymbol{f_\theta}(\boldsymbol{\vz}_{t_{n+1}}, \boldsymbol{c}, t_{n+1})$\end{small} and \begin{small}$\boldsymbol{f_\theta}(\boldsymbol{\vz}_{t_n}, \boldsymbol{c}, t_{n})$\end{small}) are already close to each other, incurring small consistency loss and hence leading to slow convergence. To address this issues, we introduce the \textsc{skipping-step} method to considerably shorten the length of time schedule (from thousands to dozens) to achieve fast convergence while preserving generation quality. 

Consistency Models (CMs) \citep{song2023consistency} use the EDM \citep{karras2022elucidating} continuous time schedule, and the Euler, or Heun Solver as the numerical continuous PF-ODE solver. 
For LCMs, in order to adapt to the discrete-time schedule in Stable Diffusion, we utilize DDIM \citep{song2020denoising}, DPM-Solver \citep{lu2022dpm}, or DPM-Solver++ \citep{lu2022dpm++} as the ODE solver. \citep{lu2022dpm} shows that these advanced solvers can solve the PF-ODE efficiently in Eq.~\ref{eq:pf_ode_vanilla}.
% \begin{equation}
%     \boldsymbol{x}_{t_{i-1} \rightarrow t_i}=\frac{\alpha_{t_i}}{\alpha_{t_{i-1}}} \tilde{\boldsymbol{x}}_{t_{i-1}}-\alpha_{t_i} \int_{\lambda_{t_{i-1}}}^{\lambda_{t_i}} e^{-\lambda} \hat{\boldsymbol{\epsilon}}_\theta\left(\hat{\boldsymbol{x}}_\lambda, \lambda\right) \mathrm{d} \lambda .
% \end{equation}
%By leveraging these advanced diffusion solvers, we can easily
Now, we introduce the \textsc{Skipping-Step} method in Latent Consistency Distillation (LCD). Instead of ensuring consistency between adjacent time steps $t_{n+1} \rightarrow t_{n}$, LCMs aim to ensure consistency between the current time step and $k$-step away, $t_{n+k} \rightarrow t_{n}$. Note that setting $k$=1 reduces to the original schedule in \citep{song2023consistency}, leading to slow convergence, and very large $k$ may incur large approximation errors
of the ODE solvers. In our main experiments, we set $k$=20, drastically reducing the length of time schedule from thousands to dozens. Results in Sec.~\ref{sec:ablation} show the effect of various $k$ values and reveal that the \textsc{skipping-step} method is crucial in accelerating the LCD process. 
Specifically, consistency distillation loss in Eq.~\ref{eq:lcm_consistency_loss_after} is modified to ensure consistency from $t_{n+k}$ to $t_{n}$:
\begin{small}
\begin{equation}
\mathcal{L_{CD}}\left(\boldsymbol{\theta},\boldsymbol{\theta^-};\Psi\right) = \mathbb{E}_{\boldsymbol{\vz,c},\omega,n}\left[d\left(\boldsymbol{f_\theta}(\boldsymbol{\vz}_{t_{n+k}}, \omega, \boldsymbol{c}, t_{n+k})\ , \boldsymbol{f_{\theta^-}}(\boldsymbol{\hat{\vz}}^{\Psi,\omega}_{t_{n}}, \omega, \boldsymbol{c}, t_{n})\right)\right],
\label{eq:lcm_consistency_loss}
\end{equation}
\end{small}
with $\boldsymbol{\hat{\vz}}^{\Psi,\omega}_{t_{n}}$ being an
estimate of $\vz_{t_{n}}$ using numerical {\em{augmented PF-ODE}} solver $\Psi$:
\begin{small}
\begin{equation}
    \hat{\vz}^{\Psi, \omega}_{t_n}\longleftarrow \vz_{t_{n+k}} + (1+\omega)\Psi(\vz_{t_{n+k}},t_{n+k}, t_n, \vc)-\omega\Psi(\vz_{t_{n+k}},t_{n+k}, t_n, \varnothing). \label{eq:aug_pf_ode_final}
\end{equation}
\end{small}
% \jian{need some details of $\Psi(\vz_{t_{n+k}},t_{n+k}, t_n, \vc)$.}
The above derivation is similar to Eq.~\ref{eq:aug_pf_ode_estimate}. For LCM, we use three possible ODE solvers here: DDIM \citep{song2020denoising}, DPM-Solver \citep{lu2022dpm}, DPM-Solver++ \citep{lu2022dpm++}, and we compare their performance
in Sec~\ref{sec:ablation}.
In fact, DDIM \citep{song2020denoising} is the first-order discretization approximation of the DPM-Solver (Proven in \citep{lu2022dpm}).
% with the exact solution \citep{lu2022dpm} of the diffusion ODE in Eq.~\ref{eq:pf_ode} from $t_{i-1} \rightarrow t_i$. 
% \jian{revise tomorrow.}
Here we provide the detailed formula of the DDIM PF-ODE solver $\Psi_{\text{DDIM}}$ from $t_{n+k}$ to $t_n$. The formulas of the other two solver $\Psi_{\text{DPM-Solver}}$, $\Psi_{\text{DPM-Solver++}}$ are provided in Appendix~\ref{appendix:solvers}.
\begin{footnotesize}
\begin{equation}
    \begin{aligned}
    \Psi_{\text{DDIM}}(\vz_{t_{n+k}},t_{n+k}, t_n, \vc) = \underbrace{\frac{\alpha_{t_n}}{\alpha_{t_{n+k}}}\vz_{t_{n+k}} - \sigma_{t_n}\left(\frac{\sigma_{t_{n+k}}\cdot\alpha_{t_n}}{\alpha_{t_{n+k}}\cdot\sigma_{t_n}} - 1\right)\hat{\boldsymbol{\epsilon}}_\theta(\vz_{t_{n+k}},\vc,t_{n+k})}_{\text{DDIM Estimated} \ \vz_{t_{n}}} - \vz_{t_{n+k}}
    % \hat{\vz}^{\Psi, w}_{t_n}\longleftarrow (1+w)\Psi(\vz_{t_{n+k}},t_{n+k}, t_n, \vc)-w\Psi(\vz_{t_{n+k}},t_{n+k}, t_n, \varnothing)
    \end{aligned}
\end{equation}
\end{footnotesize}
% \begin{footnotesize}
% \begin{equation}
%     \begin{aligned}
%     \Psi_{\text{DDIM}}(\vz_{t_{n+k}},t_{n+k}, t_n, \vc) & = \vz_{t_{n}} - \vz_{t_{n+k}}  \\
%     & = \underbrace{\frac{\alpha_{t_n}}{\alpha_{t_{n+k}}}\vz_{t_{n+k}} - \sigma_{t_n}\left(\frac{\sigma_{t_{n+k}}\cdot\alpha_{t_n}}{\alpha_{t_{n+k}}\cdot\sigma_{t_n}} - 1\right)\hat{\boldsymbol{\epsilon}}_\theta(\vz_{t_{n+k}},\vc,t_{n+k})}_{\text{DDIM Estimated} \ \vz_{t_{n}}} - \vz_{t_{n+k}}
%     % \hat{\vz}^{\Psi, w}_{t_n}\longleftarrow (1+w)\Psi(\vz_{t_{n+k}},t_{n+k}, t_n, \vc)-w\Psi(\vz_{t_{n+k}},t_{n+k}, t_n, \varnothing)
%     \end{aligned}
% \end{equation}
% \end{footnotesize}
We present the pseudo-code for \textbf{LCD}
with CFG and \textsc{skipping-step} techniques in Algorithm~\ref{alg:LCD}
The modifications from the original Consistency Distillation (CD) algorithm in \cite{song2023consistency} are highlighted in blue.  Also, the LCM sampling algorithm~\ref{alg:LCM_sample} is provided in Appendix~\ref{appendix:sampling}.
\vspace{-10pt}
\begin{algorithm}[H]
    \begin{minipage}{\linewidth}
    \begin{footnotesize}
        \caption{Latent Consistency Distillation (LCD)}\label{alg:LCD}
        \begin{algorithmic}
            \STATE \textbf{Input:} dataset $\mathcal{D}$, initial model parameter $\vtheta$, learning rate $\eta$, \textcolor{blue}{ODE solver $\Psi(\cdot,\cdot,\cdot, \cdot)$}, distance metric $d(\cdot,\cdot)$, EMA rate $\mu$, \textcolor{blue}{noise schedule $\alpha(t),\sigma(t)$, guidance scale $[w_{\text{min}},w_{\text{max}}]$, skipping interval $k$, and encoder $E(\cdot)$}
            \STATE \textcolor{blue}{Encoding training data into latent space: $\mathcal{D}_z=\{(\vz,\vc)|\vz=E(\vx),(\vx,\vc)\in\mathcal{D}\}$}
            \STATE $\vtheta^-\leftarrow\vtheta$
            \REPEAT
            \STATE Sample $(\vz,\vc) \sim \mathcal{D}_z$, \textcolor{blue}{$n\sim \mathcal{U}[1,N-k]$ and $\omega\sim [\omega_\text{min},\omega_\text{max}]$}
            \STATE Sample $\vz_{t_{n+k}}\sim \mathcal{N}(\alpha(t_{n+k})\vz;\sigma^2(t_{n+k})\mathbf{I})$
            \STATE $\begin{aligned}\textcolor{blue}{\hat{\vz}^{\Psi,\omega}_{t_n}\leftarrow \vz_{t_{n+k}}+(1+\omega)\Psi(\vz_{t_{n+k}},t_{n+k},t_n,\vc)-\omega\Psi(\vz_{t_{n+k}},t_{n+k},t_n,\varnothing)}\end{aligned}$
            \STATE $\begin{aligned}\mathcal{L}(\vtheta,\vtheta^-; \Psi)\leftarrow \textcolor{blue}{d(\vf_\vtheta(\vz_{t_{n+k}},\omega,\vc, t_{n+k}),\vf_{\vtheta^-}(\hat{\vz}^{\Psi,\omega}_{t_n},\omega,\vc, t_n))}\end{aligned}$
          \STATE $\vtheta\leftarrow\vtheta-\eta\nabla_\vtheta\mathcal{L}(\vtheta,\vtheta^-)$
            \STATE $\vtheta^-\leftarrow \text{stopgrad}(\mu\vtheta^-+(1-\mu)\vtheta)$
            \UNTIL convergence
        \end{algorithmic} 
        \end{footnotesize}
    \end{minipage}
\end{algorithm}
\vspace{-16pt}

\vspace{-10pt}
\subsection{Latent Consistency Fine-tuning for customized dataset \label{sec:consitency_finetune}}
\vspace{-8pt}
%Foundation generative models like Stable Diffusion excel in diverse text-to-image generation tasks but often require downstream finetuning on customized datasets to satisfy practical needs. We introduce Latent Consistency Fine-tuning (\textbf{LCF}), a new fine-tuning paradigm for pretrained LCM, inspired by Consistency Training (CT) in \citep{song2023consistency}, to support few-step inference on customized datasets. The LCF does not require a teacher diffusion model that is trained on a customized dataset, serving as a viable alternative to conventional diffusion model fine-tuning methods. Pseudo-code of LCF is provided in Algorithm~\ref{alg:LCF}. Detailed illustration of this algorithm is provided in \lsm{Appendix}.

Foundation generative models like Stable Diffusion excel in diverse text-to-image generation tasks but often require fine-tuning on customized datasets to meet the requirements of downstream tasks. We propose Latent Consistency Fine-tuning (LCF), a fine-tuning method for pretrained LCM. Inspired by Consistency Training (CT) \citep{song2023consistency}, LCF enables efficient few-step inference on customized datasets without relying on a teacher diffusion model trained on such data. This approach presents a viable alternative to traditional fine-tuning methods for diffusion models. 
The pseudo-code for LCF is provided in Algorithm~\ref{alg:LCF}, with a more detailed illustration in Appendix~\ref{appendix:LCF}.

\vspace{-10pt}
\section{Experiment}
\vspace{-8pt}
In this section, we employ latency consistency distillation to train LCM on two subsets of LAION-5B.
In Sec~\ref{sec:text-to-image}, we first evaluate the performance of LCM on text-to-image generation tasks. In Sec~\ref{sec:ablation}, we provide a detailed ablation study to test the effectiveness of using different solvers, skipping step schedules and guidance scales. Lastly, in Sec~\ref{sec:lcf}, we present the experimental results of latent consistency finetuning on a pretrained LCM on customized image datasets.

\vspace{-10pt}
\subsection{Text-to-Image Generation \label{sec:text-to-image}}
\vspace{-5pt}

\begin{table}[t]
\centering
\footnotesize
\setlength{\tabcolsep}{1.1pt}{
\begin{tabular}{l|cccc|cccc}
\toprule
\multicolumn{1}{c|}{\multirow{2}{*}{\textsc{Model (512 $\times$ 512) Reso}}} & \multicolumn{4}{c|}{\textsc{FID $\downarrow$}} & \multicolumn{4}{c}{\textsc{CLIP Score $\uparrow$}}      \\ 
\cline{2-9}
\multicolumn{1}{c|}{}             & \textsc{1 Step}       & \textsc{2 Steps} & \textsc{4 Steps} & \textsc{8 Steps}      & \textsc{1 Steps} & \textsc{2 Steps} & \textsc{4 Steps} & \textsc{8 Steps}   \\ 
\hline
\hline
DDIM \citep{song2020denoising}                & 183.29  & 81.05  & 22.38  & 13.83    & 6.03  & 14.13  & 25.89  & 29.29   \\
DPM \citep{lu2022dpm}                         & 185.78  & 72.81  & 18.53  & 12.24    & 6.35  & 15.10  & 26.64  & 29.54   \\
DPM++ \citep{lu2022dpm++}                     & 185.78  & 72.81  & 18.43  & 12.20    & 6.35  & 15.10  & 26.64  & \bf{29.55}   \\
Guided-Distill \citep{meng2023distillation}  & 108.21  &  33.25   & 15.12  & 13.89  & 12.08  & 22.71  & 27.25  &  28.17  \\
\textsc{LCM} (Ours)                           & \bf{35.36}  & \bf{13.31} & \bf{11.10}  & \bf{11.84} & \bf{24.14} & \bf{27.83}  & \bf{28.69}  & 28.84 \\
\bottomrule
\end{tabular}}
\vspace{-0.14in}
\caption{\footnotesize Quantitative results with $\omega=8$ at 512$\times$512 resolution. LCM significantly surpasses baselines in the 1-4 step region on LAION-Aesthetic-6+ dataset. For LCM, DDIM-Solver is used with a skipping step of $k=20$. \label{tab:512_results}}
\vspace{-8pt}
\end{table}

\begin{table}[t]
\centering
\footnotesize
\setlength{\tabcolsep}{1.1pt}{
\begin{tabular}{l|cccc|cccc}
\toprule
\multicolumn{1}{c|}{\multirow{2}{*}{\textsc{Model (768 $\times$ 768) Reso}}} & \multicolumn{4}{c|}{\textsc{FID $\downarrow$}} & \multicolumn{4}{c}{\textsc{CLIP Score $\uparrow$}}      \\ 
\cline{2-9}
\multicolumn{1}{c|}{}             & \textsc{1 Step}       & \textsc{2 Steps} & \textsc{4 Steps} & \textsc{8 Steps}      & \textsc{1 Steps} & \textsc{2 Steps} & \textsc{4 Steps} & \textsc{8 Steps}   \\ 
\hline
\hline
DDIM \citep{song2020denoising}                & 186.83  & 77.26  & 24.28  & 15.66    & 6.93  & 16.32  & 26.48  & 29.49  \\
DPM \citep{lu2022dpm}                         & 188.92  & 67.14  & 20.11  & \bf{14.08}    & 7.40  & 17.11  & 27.25  & 29.80  \\
DPM++ \citep{lu2022dpm++}                     & 188.91  & 67.14  & 20.08  & 14.11    & 7.41  & 17.11  & 27.26  & \bf{29.84}  \\
Guided-Distill \citep{meng2023distillation}  & 120.28        & 30.70  & 16.70  & 14.12   &  12.88 &  24.88 &  28.45 &  29.16  \\
\textsc{LCM} (Ours)                           & \bf{34.22}  & \bf{16.32} & \bf{13.53}  & 14.97 & \bf{25.32} & \bf{27.92}  & \bf{28.60}  & 28.49  \\
\bottomrule
\end{tabular}}
\vspace{-0.14in}
\caption{\footnotesize  Quantitative results with $\omega=8$ at 768$\times$768 resolution. LCM significantly surpasses the baselines in the 1-4 step region on LAION-Aesthetic-6.5+ dataset. For LCM, DDIM-Solver is used with a skipping step of $k=20$. \label{tab:768_results}}
\vspace{-15pt}
\end{table}

\noindent \textbf{Datasets} We use two subsets of LAION-5B \citep{schuhmann2022laion}: LAION-Aesthetics-6+ (12M) and LAION-Aesthetics-6.5+ (650K) for text-to-image generation. Our experiments consider resolutions of 512$\times$512 and 768$\times$768. For 512 resolution, we use LAION-Aesthetics-6+, which comprises 12M text-image pairs with predicted aesthetics scores higher than 6. For 768 resolution, we use LAION-Aesthetics-6.5+, with 650K text-image pairs with aesthetics score higher than 6.5. 

\noindent \textbf{Model Configuration} For 512 resolution, we use the pre-trained Stable Diffusion-V2.1-Base \citep{rombach2022high} as the teacher model, which was originally trained on resolution 512$\times$512 with $\boldsymbol{\epsilon}$-Prediction \citep{ho2020denoising}. For 768 resolution, we use the widely used pre-trained Stable Diffusion-V2.1, originally trained on resolution 768$\times$768 with $\boldsymbol{v}$-Prediction \citep{salimans2022progressive}. We train LCM with 100K iterations and we use a batch size of 72 for $(512\times512)$ setting, and 16 for $(768\times768)$ setting, the same learning rate 8e-6 and EMA rate $\mu=0.999943$ as used in \citep{song2023consistency}. For {\em augmented PF-ODE} solver $\Psi$ and skipping step $k$ in Eq.~\ref{eq:aug_pf_ode_final}, we use DDIM-Solver \citep{song2020denoising} with skipping step $k=20$. We set the guidance scale range $[w_{\text{min}},w_{\text{max}}] = [2, 14]$, consistent with \citep{meng2023distillation}. More training details are provided in the Appendix~\ref{appendix:training_details}.

\noindent \textbf{Baselines \& Evaluation} We use DDIM \citep{song2020denoising}, DPM \citep{lu2022dpm}, DPM++ \citep{lu2022dpm++} and Guided-Distill \citep{meng2023distillation} as baselines. The first three are training-free samplers requiring more peak memory per step with classifier-free guidance. Guided-Distill requires two stages of guided distillation. Since Guided-Distill is not open-sourced, we strictly followed the training procedure outlined in the paper to reproduce the results. Due to the limited resource (\cite{meng2023distillation} used a large batch size of 512, requiring at least 32 A100 GPUs), we reduce the batch size to $72$, the same as ours, and trained for the same 100K iterations. Reproduction details are provided in Appendix~\ref{appendix:reproduction}. We admit that longer training and more computational resources can lead to better results as reported in \citep{meng2023distillation}. However, LCM achieves faster convergence and superior results under the same computation cost. For evaluation, 
We generate 30K images from 10K text prompts in the test set (3 images per prompt), and
adopt FID and CLIP scores to evaluate the diversity and quality
of the generated images. We use ViT-g/14 for evaluating CLIP scores. 

\begin{figure}[t] 
\begin{centering}
\includegraphics[scale=0.48]{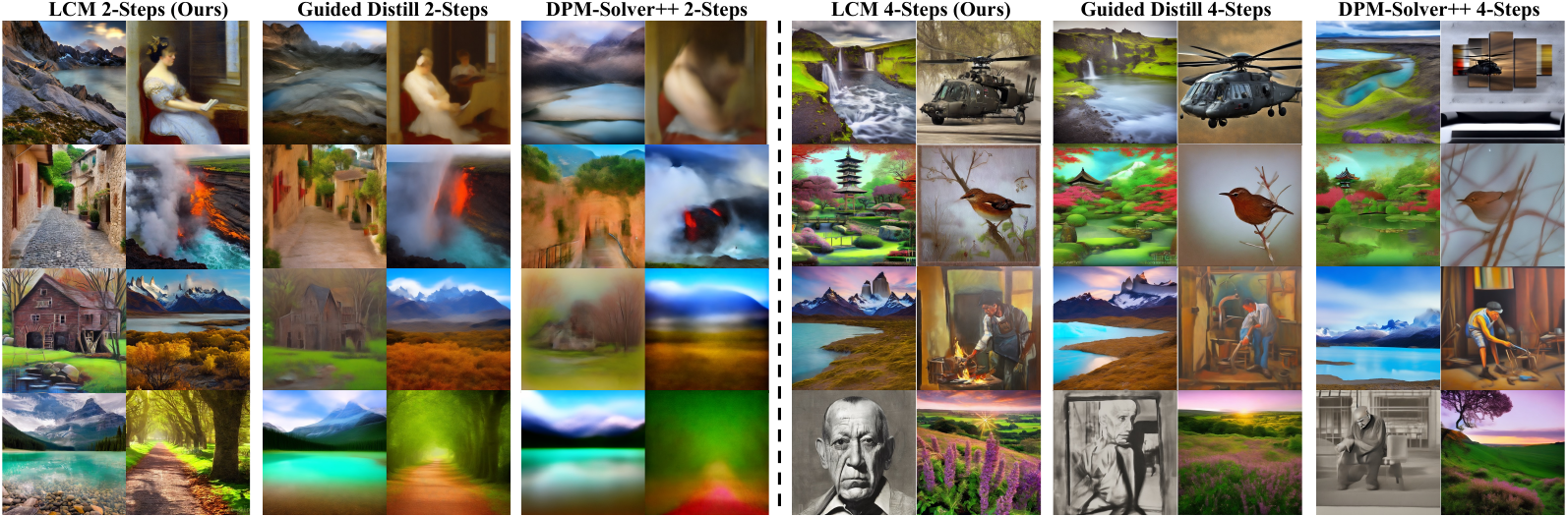}
 \vspace{-0.14in}
\caption{\footnotesize{Text-to-Image generation results on LAION-Aesthetic-6.5+ with 2-, 4-step inference. Images generated by LCM exhibit superior detail and quality, outperforming other baselines by a large margin.}\label{fig:text_to_image}}
\end{centering}
\vspace{-0.14in}
\end{figure}

\noindent \textbf{Results.} The quantitative results in Tables~\ref{tab:512_results} and ~\ref{tab:768_results} show that LCM notably outperforms baseline methods at $512$ and $768$ resolutions, especially in the low step regime (1$\sim$4), highlighting its efficency and superior performance. Unlike DDIM, DPM, DPM++, which require more peak memory per sampling step with CFG, LCM requires only one forward pass per sampling step, saving both time and memory. Moreover, in contrast to the two-stage distillation procedure employed in Guided-Distill, LCM only needs one-stage guided distillation, which is much simpler and more practical. 
The \textbf{qualitative results} in Figure~\ref{fig:text_to_image} further show
the superiority of LCM with 2- and 4-step inference.

\begin{figure}[t] 
\begin{centering}
\includegraphics[scale=0.93]{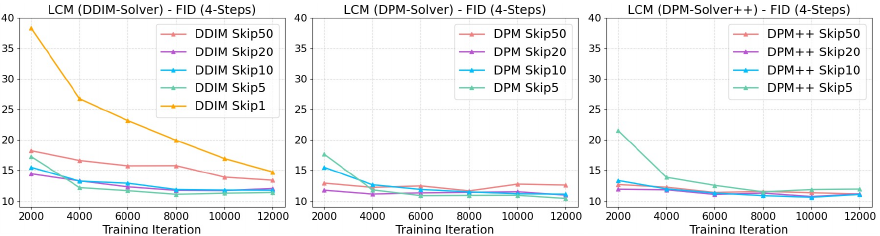}
 \vspace{-0.12in}
\caption{\footnotesize{Ablation study on different ODE solvers and skipping step $k$. Appropriate skipping step $k$ can significantly accelerate convergence and lead to better FID within the same number of training steps.}
\label{fig:ablation_solver_skip}}
\vspace{-0.16in}
\end{centering}
\end{figure}

\vspace{-10pt}
\begin{figure}[h] 
\begin{centering}
\includegraphics[scale=0.70]{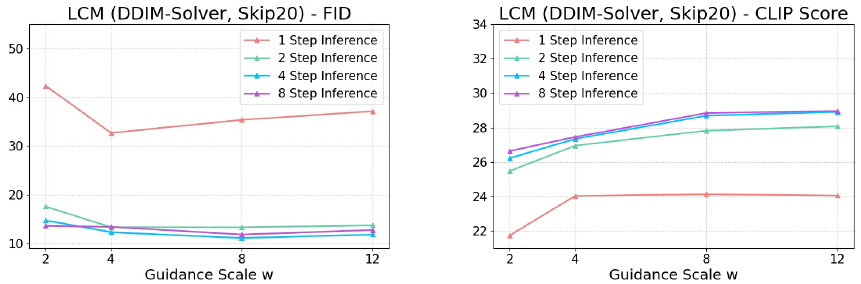}
 \vspace{-12pt}
\caption{\footnotesize{Ablation study on different classifier-free guidance scales $\omega$. Larger $\omega$ leads to better sample quality (CLIP Scores). The performance gaps across 2, 4, and 8 steps are minimal, showing the efficacy of LCM.}\label{fig:ablation_cfg_scale}}
\vspace{-8pt}
\end{centering}
% \vspace{-0.25in}
\end{figure}

\vspace{-10pt}
\subsection{Ablation Study \label{sec:ablation}}
\vspace{-8pt}
\begin{figure}[t] 
\begin{centering}
\includegraphics[scale=0.7]{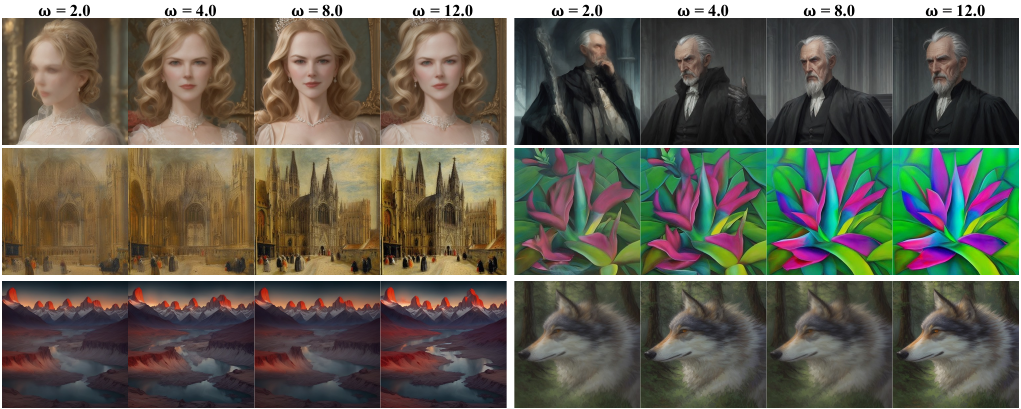}
 \vspace{-10pt}
\caption{\footnotesize{4-step LCMs using different CFG scales $\omega$. LCMs utilize one-stage guided distillation to directly incorporate CFG scales $\omega$. Larger $\omega$ enhances image quality.}
\label{fig:cfg_image}}
\vspace{-11pt}
\end{centering}
% \vspace{-0.22in}
\end{figure}

\begin{figure}[t] 
\begin{centering} 
\includegraphics[scale=0.7]{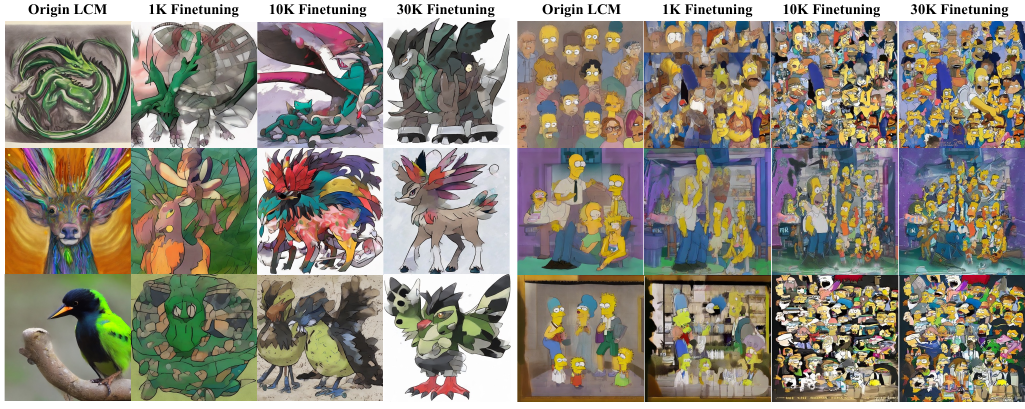}
 \vspace{-10pt}
\caption{\footnotesize{4-step LCMs using Latent Consistency Fine-tuning (LCF) on two customized datasets: Pokemon Dataset (left), Simpsons Dataset (right). Through LCF, LCM produces images with customized styles.}\label{fig:lcf}}
\end{centering}
\vspace{-15pt}
\end{figure}

\noindent \textbf{ODE Solvers \& Skipping-Step Schedule.} We compare various solvers $\Psi$ (DDIM \citep{song2020denoising}, DPM \citep{lu2022dpm}, DPM++ \citep{lu2022dpm++}) for 
solving the {\em{augmented PF-ODE}} specified in Eq~\ref{eq:aug_pf_ode_final}, and explore different skipping step schedules with different $k$. The results are depicted in Figure~\ref{fig:ablation_solver_skip}. We observe that: 1) Using \textsc{Skipping-Step} techniques (see Sec~\ref{sec:skipping_step}), LCM achieves fast convergence within 2,000 iterations in the 4-step inference setting. Specifically, the DDIM solver converges slowly at skipping step $k=1$, while setting $k=5,10,20$ leads to much faster convergence, underscoring the effectiveness of the Skipping-Step method. 2) DPM and DPM++ solvers perform better at a larger skipping step ($k=50$) compared to the DDIM solver which suffers from increased ODE approximation error with larger $k$. This phenomenon is also discussed in \citep{lu2022dpm}. 3) Very small $k$ values (1 or 5) result in slow convergence and very large ones (e.g., 50 for DDIM) may lead to inferior results. Hence, we choose $k=20$, which provides competitive performance for all three solvers, for our main experiment in Sec~\ref{sec:text-to-image}.

\noindent \textbf{The Effect of Guidance Scale $\omega$.} We examine the effect of using different CFG scales $\omega$ in LCM. Typically, $\omega$ balances sample quality and diversity. A larger $\omega$ generally tends to improve sample quality (indicated by CLIP), but may compromise diversity (measured by FID). Beyond a certain threshold, an increased $\omega$ yields better CLIP scores at the expense of FID. 
Figure~\ref{fig:ablation_cfg_scale} presents the results for various $\omega$ across different inference steps.
%Results for various $\omega$ across different inference steps are shown in Figure~\ref{fig:ablation_cfg_scale}. 
Our findings include: 1) Using large $\omega$ enhances sample quality (CLIP Scores) but results in relatively inferior FID. 2) The performance gaps across 2, 4, and 8 inference steps are negligible, highlighting LCM's efficacy in 2$\sim$8 step regions. However, a noticeable gap exists in one-step inference, indicating rooms for further improvements. We present visualizations for different $\omega$ in Figure~\ref{fig:cfg_image}. One can see clearly that a larger $\omega$ enhances sample quality, verifying the effectiveness of our one-stage guided distillation method.

\vspace{-10pt}
\subsection{Downstream Consistency Fine-tuning Results \label{sec:lcf}}
\vspace{-5pt}
We perform \textbf{Latent Consistency Fine-tuning} (LCF) on two customized image datasets, Pokemon dataset \citep{pinkney2022pokemon} and Simpsons dataset \citep{norod2022simpsons}, to demonstrate the efficiency of LCF.
%the potential of LCF as a new fine-tuning method to support few-step inference on customized datasets by finetuning a pretrained LCM.
Each dataset, comprised of hundreds of customized text-image pairs, is split such that 90\% is used for fine-tuning and the rest 10\% for testing. For LCF, we utilize pretrained LCM that was originally trained at the resolution of 768$\times$768 used in Table~\ref{tab:768_results}. For these two datasets, we fine-tune the pre-trained LCM for 30K iterations with a learning rate 8e-6. We present qualitative results of adopting LCF on two customized image datasets in Figure~\ref{fig:lcf}. The finetuned LCM is capable of generating images with customized styles in few steps, showing the effectiveness of our method.

\vspace{-12pt}
\section{Conclusion}
\vspace{-8pt}
We present Latent Consistency Models (LCMs), and a highly efficient one-stage guided distillation method that enables few-step or even one-step inference on pre-trained LDMs.
%Remarkably, LCM requires only 4K training iterations to support 2$\sim$4-steps inference and even one-step inference, a significant efficiency improvement over existing guided distillation methods.
Furthermore, we present latent consistency fine-tuning (LCF), to enable few-step inference of LCMs on customized image datasets. Extensive experiments on the LAION-5B-Aesthetics dataset demonstrate the superior performance and efficiency of LCMs.
Future work include extending our method to more image generation tasks such as 
text-guided image editing, inpainting and super-resolution.

% \section*{Reproducibility Statement}
% In our paper, we discuss the data, model, training hyper-parameters as detailed in Section~\ref{sec:text-to-image}, Appendix~\ref{appendix:training_details}. Since our approach is straightforward and computation efficient, it ensures a high level of reproducibility of our work.

\bibliography{iclr2024_conference}

\begin{thebibliography}{35}
\providecommand{\natexlab}[1]{#1}
\providecommand{\url}[1]{\texttt{#1}}
\expandafter\ifx\csname urlstyle\endcsname\relax
  \providecommand{\doi}[1]{doi: #1}\else
  \providecommand{\doi}{doi: \begingroup \urlstyle{rm}\Url}\fi

\bibitem[Deng et~al.(2009)Deng, Dong, Socher, Li, Li, and Fei-Fei]{deng2009imagenet}
Jia Deng, Wei Dong, Richard Socher, Li-Jia Li, Kai Li, and Li~Fei-Fei.
\newblock Imagenet: A large-scale hierarchical image database.
\newblock In \emph{2009 IEEE conference on computer vision and pattern recognition}, pp.\  248--255. Ieee, 2009.

\bibitem[Goodfellow et~al.(2020)Goodfellow, Pouget-Abadie, Mirza, Xu, Warde-Farley, Ozair, Courville, and Bengio]{goodfellow2020generative}
Ian Goodfellow, Jean Pouget-Abadie, Mehdi Mirza, Bing Xu, David Warde-Farley, Sherjil Ozair, Aaron Courville, and Yoshua Bengio.
\newblock Generative adversarial networks.
\newblock \emph{Communications of the ACM}, 63\penalty0 (11):\penalty0 139--144, 2020.

\bibitem[Ho \& Salimans(2022)Ho and Salimans]{ho2022classifier}
Jonathan Ho and Tim Salimans.
\newblock Classifier-free diffusion guidance.
\newblock \emph{arXiv preprint arXiv:2207.12598}, 2022.

\bibitem[Ho et~al.(2020)Ho, Jain, and Abbeel]{ho2020denoising}
Jonathan Ho, Ajay Jain, and Pieter Abbeel.
\newblock Denoising diffusion probabilistic models.
\newblock \emph{Advances in Neural Information Processing Systems}, 33:\penalty0 6840--6851, 2020.

\bibitem[Hyv{\"a}rinen \& Dayan(2005)Hyv{\"a}rinen and Dayan]{hyvarinen2005estimation}
Aapo Hyv{\"a}rinen and Peter Dayan.
\newblock Estimation of non-normalized statistical models by score matching.
\newblock \emph{Journal of Machine Learning Research}, 6\penalty0 (4), 2005.

\bibitem[Jolicoeur-Martineau et~al.(2021)Jolicoeur-Martineau, Li, Pich{\'e}-Taillefer, Kachman, and Mitliagkas]{jolicoeur2021gotta}
Alexia Jolicoeur-Martineau, Ke~Li, R{\'e}mi Pich{\'e}-Taillefer, Tal Kachman, and Ioannis Mitliagkas.
\newblock Gotta go fast when generating data with score-based models.
\newblock \emph{arXiv preprint arXiv:2105.14080}, 2021.

\bibitem[Karras et~al.(2022)Karras, Aittala, Aila, and Laine]{karras2022elucidating}
Tero Karras, Miika Aittala, Timo Aila, and Samuli Laine.
\newblock Elucidating the design space of diffusion-based generative models.
\newblock \emph{Advances in Neural Information Processing Systems}, 35:\penalty0 26565--26577, 2022.

\bibitem[Kingma \& Welling(2013)Kingma and Welling]{kingma2013auto}
Diederik~P Kingma and Max Welling.
\newblock Auto-encoding variational bayes.
\newblock \emph{arXiv preprint arXiv:1312.6114}, 2013.

\bibitem[Liu et~al.(2019)Liu, Jiang, He, Chen, Liu, Gao, and Han]{liu2019variance}
Liyuan Liu, Haoming Jiang, Pengcheng He, Weizhu Chen, Xiaodong Liu, Jianfeng Gao, and Jiawei Han.
\newblock On the variance of the adaptive learning rate and beyond.
\newblock \emph{arXiv preprint arXiv:1908.03265}, 2019.

\bibitem[Liu et~al.(2022)Liu, Gong, and Liu]{liu2022flow}
Xingchao Liu, Chengyue Gong, and Qiang Liu.
\newblock Flow straight and fast: Learning to generate and transfer data with rectified flow.
\newblock \emph{arXiv preprint arXiv:2209.03003}, 2022.

\bibitem[Liu et~al.(2023)Liu, Zhang, Ma, Peng, and Liu]{liu2023instaflow}
Xingchao Liu, Xiwen Zhang, Jianzhu Ma, Jian Peng, and Qiang Liu.
\newblock Instaflow: One step is enough for high-quality diffusion-based text-to-image generation.
\newblock \emph{arXiv preprint arXiv:2309.06380}, 2023.

\bibitem[Lu et~al.(2022{\natexlab{a}})Lu, Zhou, Bao, Chen, Li, and Zhu]{lu2022dpm}
Cheng Lu, Yuhao Zhou, Fan Bao, Jianfei Chen, Chongxuan Li, and Jun Zhu.
\newblock Dpm-solver: A fast ode solver for diffusion probabilistic model sampling in around 10 steps.
\newblock \emph{arXiv preprint arXiv:2206.00927}, 2022{\natexlab{a}}.

\bibitem[Lu et~al.(2022{\natexlab{b}})Lu, Zhou, Bao, Chen, Li, and Zhu]{lu2022dpm++}
Cheng Lu, Yuhao Zhou, Fan Bao, Jianfei Chen, Chongxuan Li, and Jun Zhu.
\newblock Dpm-solver++: Fast solver for guided sampling of diffusion probabilistic models.
\newblock \emph{arXiv preprint arXiv:2211.01095}, 2022{\natexlab{b}}.

\bibitem[Lyu et~al.(2022)Lyu, Xu, Yang, Lin, and Dai]{lyu2022accelerating}
Zhaoyang Lyu, Xudong Xu, Ceyuan Yang, Dahua Lin, and Bo~Dai.
\newblock Accelerating diffusion models via early stop of the diffusion process.
\newblock \emph{arXiv preprint arXiv:2205.12524}, 2022.

\bibitem[Meng et~al.(2023)Meng, Rombach, Gao, Kingma, Ermon, Ho, and Salimans]{meng2023distillation}
Chenlin Meng, Robin Rombach, Ruiqi Gao, Diederik Kingma, Stefano Ermon, Jonathan Ho, and Tim Salimans.
\newblock On distillation of guided diffusion models.
\newblock In \emph{Proceedings of the IEEE/CVF Conference on Computer Vision and Pattern Recognition}, pp.\  14297--14306, 2023.

\bibitem[Nichol et~al.(2021)Nichol, Dhariwal, Ramesh, Shyam, Mishkin, McGrew, Sutskever, and Chen]{Alex2021glide}
Alex Nichol, Prafulla Dhariwal, Aditya Ramesh, Pranav Shyam, Pamela Mishkin, Bob McGrew, Ilya Sutskever, and Mark Chen.
\newblock Glide: Towards photorealistic image generation and editing with text-guided diffusion models.
\newblock \emph{arXiv preprint arXiv:2112.10741}, 2021.

\bibitem[Nichol \& Dhariwal(2021)Nichol and Dhariwal]{nichol2021improved}
Alexander~Quinn Nichol and Prafulla Dhariwal.
\newblock Improved denoising diffusion probabilistic models.
\newblock In \emph{International Conference on Machine Learning}, pp.\  8162--8171. PMLR, 2021.

\bibitem[Norod78(2022)]{norod2022simpsons}
Norod78.
\newblock Simpsons blip captions.
\newblock \url{https://huggingface.co/datasets/Norod78/simpsons-blip-captions}, 2022.

\bibitem[Pinkney(2022)]{pinkney2022pokemon}
Justin N.~M. Pinkney.
\newblock Pokemon blip captions.
\newblock \url{https://huggingface.co/datasets/lambdalabs/pokemon-blip-captions/}, 2022.

\bibitem[Ramesh et~al.(2022)Ramesh, Dhariwal, Nichol, Chu, and Chen]{ramesh2022hierarchical}
Aditya Ramesh, Prafulla Dhariwal, Alex Nichol, Casey Chu, and Mark Chen.
\newblock Hierarchical text-conditional image generation with clip latents.
\newblock \emph{arXiv preprint arXiv:2204.06125}, 2022.

\bibitem[Rombach et~al.(2022)Rombach, Blattmann, Lorenz, Esser, and Ommer]{rombach2022high}
Robin Rombach, Andreas Blattmann, Dominik Lorenz, Patrick Esser, and Bj{\"o}rn Ommer.
\newblock High-resolution image synthesis with latent diffusion models.
\newblock In \emph{Proceedings of the IEEE/CVF Conference on Computer Vision and Pattern Recognition}, pp.\  10684--10695, 2022.

\bibitem[Saharia et~al.(2022)Saharia, Chan, Saxena, Li, Whang, Denton, Ghasemipour, Gontijo~Lopes, Karagol~Ayan, Salimans, et~al.]{saharia2022photorealistic}
Chitwan Saharia, William Chan, Saurabh Saxena, Lala Li, Jay Whang, Emily~L Denton, Kamyar Ghasemipour, Raphael Gontijo~Lopes, Burcu Karagol~Ayan, Tim Salimans, et~al.
\newblock Photorealistic text-to-image diffusion models with deep language understanding.
\newblock \emph{Advances in Neural Information Processing Systems}, 35:\penalty0 36479--36494, 2022.

\bibitem[Salimans \& Ho(2022)Salimans and Ho]{salimans2022progressive}
Tim Salimans and Jonathan Ho.
\newblock Progressive distillation for fast sampling of diffusion models.
\newblock \emph{arXiv preprint arXiv:2202.00512}, 2022.

\bibitem[Schuhmann et~al.(2022)Schuhmann, Beaumont, Vencu, Gordon, Wightman, Cherti, Coombes, Katta, Mullis, Wortsman, et~al.]{schuhmann2022laion}
Christoph Schuhmann, Romain Beaumont, Richard Vencu, Cade Gordon, Ross Wightman, Mehdi Cherti, Theo Coombes, Aarush Katta, Clayton Mullis, Mitchell Wortsman, et~al.
\newblock Laion-5b: An open large-scale dataset for training next generation image-text models.
\newblock \emph{arXiv preprint arXiv:2210.08402}, 2022.

\bibitem[Sohn et~al.(2015)Sohn, Lee, and Yan]{sohn2015learning}
Kihyuk Sohn, Honglak Lee, and Xinchen Yan.
\newblock Learning structured output representation using deep conditional generative models.
\newblock \emph{Advances in neural information processing systems}, 28, 2015.

\bibitem[Song et~al.(2020{\natexlab{a}})Song, Meng, and Ermon]{song2020denoising}
Jiaming Song, Chenlin Meng, and Stefano Ermon.
\newblock Denoising diffusion implicit models.
\newblock \emph{arXiv preprint arXiv:2010.02502}, 2020{\natexlab{a}}.

\bibitem[Song \& Ermon(2019)Song and Ermon]{song2019generative}
Yang Song and Stefano Ermon.
\newblock Generative modeling by estimating gradients of the data distribution.
\newblock \emph{Advances in Neural Information Processing Systems}, 32, 2019.

\bibitem[Song et~al.(2020{\natexlab{b}})Song, Sohl-Dickstein, Kingma, Kumar, Ermon, and Poole]{song2020score}
Yang Song, Jascha Sohl-Dickstein, Diederik~P Kingma, Abhishek Kumar, Stefano Ermon, and Ben Poole.
\newblock Score-based generative modeling through stochastic differential equations.
\newblock \emph{arXiv preprint arXiv:2011.13456}, 2020{\natexlab{b}}.

\bibitem[Song et~al.(2021)Song, Durkan, Murray, and Ermon]{song2021maximum}
Yang Song, Conor Durkan, Iain Murray, and Stefano Ermon.
\newblock Maximum likelihood training of score-based diffusion models.
\newblock \emph{Advances in Neural Information Processing Systems}, 34:\penalty0 1415--1428, 2021.

\bibitem[Song et~al.(2023)Song, Dhariwal, Chen, and Sutskever]{song2023consistency}
Yang Song, Prafulla Dhariwal, Mark Chen, and Ilya Sutskever.
\newblock Consistency models.
\newblock \emph{arXiv preprint arXiv:2303.01469}, 2023.

\bibitem[Watson et~al.(2021)Watson, Ho, Norouzi, and Chan]{watson2021learning}
Daniel Watson, Jonathan Ho, Mohammad Norouzi, and William Chan.
\newblock Learning to efficiently sample from diffusion probabilistic models.
\newblock \emph{arXiv preprint arXiv:2106.03802}, 2021.

\bibitem[Yu et~al.(2015)Yu, Seff, Zhang, Song, Funkhouser, and Xiao]{yu2015lsun}
Fisher Yu, Ari Seff, Yinda Zhang, Shuran Song, Thomas Funkhouser, and Jianxiong Xiao.
\newblock Lsun: Construction of a large-scale image dataset using deep learning with humans in the loop.
\newblock \emph{arXiv preprint arXiv:1506.03365}, 2015.

\bibitem[Zhang \& Agrawala(2023)Zhang and Agrawala]{zhang2023adding}
Lvmin Zhang and Maneesh Agrawala.
\newblock Adding conditional control to text-to-image diffusion models.
\newblock \emph{arXiv preprint arXiv:2302.05543}, 2023.

\bibitem[Zheng et~al.(2023)Zheng, Nie, Vahdat, Azizzadenesheli, and Anandkumar]{zheng2023fast}
Hongkai Zheng, Weili Nie, Arash Vahdat, Kamyar Azizzadenesheli, and Anima Anandkumar.
\newblock Fast sampling of diffusion models via operator learning.
\newblock In \emph{International Conference on Machine Learning}, pp.\  42390--42402. PMLR, 2023.

\bibitem[Zheng et~al.(2022)Zheng, He, Chen, and Zhou]{zheng2022truncated}
Huangjie Zheng, Pengcheng He, Weizhu Chen, and Mingyuan Zhou.
\newblock Truncated diffusion probabilistic models.
\newblock \emph{stat}, 1050:\penalty0 7, 2022.

\end{thebibliography}
\bibliographystyle{iclr2024_conference}

\appendix

\section{More Details on Diffusion and Consistency Models}
\label{app:preliminary}

\subsection{Diffusion Models}

Consider the forward process, described by the following SDE for $t\in[0, T]$:
\begin{small}
\begin{equation}
    \mathrm{d} \boldsymbol{x}_t=f(t) \boldsymbol{x}_t \mathrm{~d} t+g(t) \mathrm{d} \boldsymbol{w}_t, \quad \boldsymbol{x}_0 \sim p_{data}\left(\boldsymbol{x}_0\right),
\end{equation}
\end{small}
where $\boldsymbol{w}_t$ denotes the standard Brownian motion.
Leveraging the classic result of Anderson (1982), \cite{song2020score} show that the reverse process of the above forward process is also a diffusion process, specified by
the following reverse-time SDE:
\begin{small}
\begin{equation}
    \mathrm{d} \boldsymbol{x}_t=\left[f(t) \boldsymbol{x}_t-g^2(t) \nabla_{\boldsymbol{x}} \log q_t\left(\boldsymbol{x}_t\right)\right] \mathrm{d} t+g(t) \mathrm{d} \overline{\boldsymbol{w}}_t, \quad \boldsymbol{x}_T \sim q_T\left(\boldsymbol{x}_T\right) \label{eq:backward_sde},
\end{equation}
\end{small}
where $\overline{\boldsymbol{w}}_t$ is a standard reverse-time Brownian motion.
One can leverage the reverse SDE for data sampling from $T$ to $0$, starting with $q_T(\boldsymbol{x}_T)$, which follows a Gaussian distribution approximately. 
However, directly sampling from the reverse SDE requires a large number of discretization steps and is typically very slow.
To accelerate the sampling process, prior work (e.g., \citep{song2020score, lu2022dpm} leveraged
the relation between the above SDE and ODE and designed ODE solvers for sampling.
In particular, it is known that for SDE (Eq.\ref{eq:backward_sde}), the following ordinary differential equation (ODE), called the \textit{Probability Flow ODE} (PF-ODE), has the same marginal distribution $q_t(\boldsymbol{x})$ \citep{song2020score, lu2022dpm}:
\begin{small}
\begin{equation}
    \frac{\mathrm{d} \boldsymbol{x}_t}{\mathrm{~d} t}=f(t) \boldsymbol{x}_t-\frac{1}{2} g^2(t) \nabla_{\boldsymbol{x}} \log q_t\left(\boldsymbol{x}_t\right), \quad \boldsymbol{x}_T \sim q_T\left(\boldsymbol{x}_T\right) \label{eq:pf_origin_ode}
\end{equation}
\end{small}
The term $-\nabla \log q_t(\boldsymbol{x}_t)$ in Eq.~\ref{eq:pf_origin_ode} is 
typically called the {\em score function} of $q_t(\boldsymbol{x}_t)$. In diffusion models, we train the noise prediction model $\boldsymbol{\epsilon}_\theta(\boldsymbol{x}_t, t)$ to fit the scaled score function, via minimizing the following score matching objective:
%\begin{small}
\begin{align}
    \mathcal{L}(\theta) & = \mathbb{E}_{t\in [0,T], x_t\sim q_t} 
    \left[w(t)||\boldsymbol{\epsilon}_\theta(\boldsymbol{x}_t, t) 
    + \sigma(t)\nabla \log q_t(\boldsymbol{x}_t))||^2\right]  \notag\\
    & =\mathbb{E}_{t\in [0,T], \boldsymbol{x}_0\sim q_0, \boldsymbol{\epsilon} } \left[w(t)||\boldsymbol{\epsilon}_\theta(\boldsymbol{x}_t, t)- 
    \boldsymbol{\epsilon}||^2\right]  
    \label{eq:score_matching}
\end{align}
%\end{small}
where $w(t)$ is the weight function, $\boldsymbol{\epsilon}\sim N(0, I)$
and $\boldsymbol{x}_t=\alpha(t) \boldsymbol{x}_0+\sigma(t) \boldsymbol{\epsilon}$.
By substituting the score function with the noise prediction model in Eq.~\ref{eq:pf_origin_ode}, we obtain the following ODE, which can be used for sampling:
\begin{small}
\begin{equation}
    \frac{\mathrm{d} \boldsymbol{x}_t}{\mathrm{~d} t}= f(t) \boldsymbol{x}_t+\frac{g^2(t)}{2 \sigma_t} \boldsymbol{\epsilon}_\theta\left(\boldsymbol{x}_t, t\right), \quad \boldsymbol{x}_T \sim \mathcal{N}\left(\mathbf{0}, \tilde{\sigma}^2 \boldsymbol{I}\right).
    \label{eq:pf_empirical_ode}
\end{equation}
\end{small}

\subsection{More Details on Consistency Models in \citep{song2023consistency}}

In this subsection, we provide more details on the consistency models and consistency distillation algorithm in \citep{song2023consistency}.
The pre-trained diffusion model used in \citep{song2023consistency}
adopts the continuous noise schedule from EDM \citep{karras2022elucidating}, therefore the PF-ODE in Eq.~\ref{eq:pf_empirical_ode} can be simplified as:
\begin{small}
\begin{equation}
    \frac{\mathrm{d} \mathbf{x}_t}{\mathrm{~d} t}
    =-t \nabla \log q_t(\boldsymbol{x}_t)
    \approx -t \boldsymbol{s}_\phi\left(\mathbf{x}_t, t\right), \label{eq:simple_pf_ode}
\end{equation}
\end{small}
where the $\boldsymbol{s}_\phi\left(\mathbf{x}_t, t\right)\approx 
\nabla \log q_t(\boldsymbol{x}_t)$ is a score prediction model trained via score matching \citep{hyvarinen2005estimation,song2019generative}. 
Note that different noise schedules result in different PF-ODE and
the PF-ODE in Eq.~\ref{eq:simple_pf_ode} corresponds to the EDM noise schedule \citep{karras2022elucidating}. 
We denote the one-step ODE solver applied to PF-ODE in Eq.~\ref{eq:simple_pf_ode} as $\Phi(\vx_t,t;\phi)$. 
One can either use Euler \citep{song2020score} or Heun solver \citep{karras2022elucidating} as the numerical ODE solver.
Then, we use the ODE solver to estimate the evolution of a sample $\boldsymbol{x}_{t_n}$ from $\boldsymbol{x}_{t_{n+1}}$ as:
\begin{equation}
    \hat{\vx}_{t_n}^\phi\leftarrow \vx_{t_{n+1}}+(t_n-t_{n+1})\Phi(\vx_{t_{n+1}},t_{n+1};\phi).
\end{equation}
\citep{song2020score} used the same time schedule as in \citep{karras2022elucidating}: 
\begin{footnotesize}$t_i=(\epsilon^{1 / \rho}+ \frac{i-1}{N-1}(T^{1 / \rho}-\epsilon^{1 / \rho}))^\rho$\end{footnotesize}, and $\rho=7$.
 % \jian{need to specify the formula for $t_i$.}
To enforce the self-consistency property in Eq.~\ref{eq:self_consistency},
we maintain a target model $\vtheta^-$, which is updated with exponential moving average (EMA) of the parameter $\vtheta$ we intend to learn, i.e., $\vtheta^-\leftarrow\mu\vtheta^-+(1-\mu)\vtheta$,
% \begin{small}
% \begin{equation}
%     \vtheta^-\leftarrow\mu\vtheta^-+(1-\mu)\vtheta. \label{eq:ema}
% \end{equation}
% \end{small}
and define the consistency loss as follows:
\begin{small}
\begin{equation}
    \mathcal{L}(\vtheta,\vtheta^-;\Phi)=\mathbb{E}_{\boldsymbol{x},t}
    \left[d\left(\vf_\vtheta(\vx_{t_{n+1}},t_{n+1}),\vf_{\vtheta^{-}}(\hat{\vx}^\phi_{t_n},t_n)\right)\right],\label{eq:consistency_loss}
\end{equation}
\end{small}
where $d(\cdot,\cdot)$ is a chosen metric function for measuring the distance between two samples, e.g., the squared $\ell_2$ distance \begin{small}$d(\vx,\vy)=||\vx-\vy||_2^2$\end{small}. The pseudo-code for consistency distillation in \cite{song2023consistency}.
is presented in Algorithm~\ref{alg:CD_raw}.
% The parameter $\vtheta$ is learnt using stochastic gradient descent until convergence. The pseudo-code is presented in Algorithm~\ref{alg:CD_raw} in Appendix~\ref{xxx}.
%Algorithm~\ref{alg:CD_raw} presents pseudo-code for achieving consistency distillation in \cite{song2023consistency}. 
In their original paper, an Euler solver was used as the ODE solver for the continuous-time setting.

\begin{algorithm}[H]
		\begin{minipage}{\linewidth}
			\caption{Consistency Distillation (CD) \citep{song2023consistency}}\label{alg:CD_raw}
			\begin{algorithmic}
                \STATE \textbf{Input:} dataset $\mathcal{D}$, initial model parameter $\vtheta$, learning rate $\eta$, ODE solver $\Phi(\cdot,\cdot,\cdot)$, distance metric $d(\cdot,\cdot)$, and EMA rate $\mu$
				\STATE $\vtheta^-\leftarrow\vtheta$
                \REPEAT
				\STATE Sample $\vx \sim \mathcal{D}$ and $n\sim \mathcal{U}[1,N-1]$
				\STATE Sample $\vx_{t_{n+1}}\sim \mathcal{N}(\vx;t^2_{n+1}\mathbf{I})$
                \STATE $\begin{aligned}\hat{\vx}^\phi_{t_n}\leftarrow \vx_{t_{n+1}}+(t_n-t_{n+1})\Phi(\vx_{t_{n+1}},t_{n+1},\phi)\end{aligned}$
                \STATE $\begin{aligned}\mathcal{L}(\vtheta,&\vtheta^-; \Phi)\leftarrow d(\vf_\vtheta(\vx_{t_{n+1}},t_{n+1}),\vf_{\vtheta^-}(\hat{\vx}_{t_n}^\phi,t_n))\end{aligned}$
		      \STATE $\vtheta\leftarrow\vtheta-\eta\nabla_\vtheta\mathcal{L}(\vtheta,\vtheta^-; \Phi)$
                \STATE $\vtheta^-\leftarrow \text{stopgrad}(\mu\vtheta^-+(1-\mu)\vtheta)$
                \UNTIL convergence
			\end{algorithmic} 
		\end{minipage}
	\end{algorithm}

\section{Multistep Latent Consistency Sampling \label{appendix:sampling}}

Now, we present the multi-step sampling algorithm for latent consistency model. 
The sampling algorithm for LCM is very similar to the one in consistency models \citep{song2023consistency} except the incorporation of classifier-free guidance in LCM.
Unlike multi-step sampling in diffusion models, in which we predict $\vz_{t-1}$ from $\vz_{t}$, 
the latent consistency models directly predicts the origin $\vz_0$ of augmented PF-ODE trajectory (the solution of the augmented of PF-ODE), given guidance scale $\omega$. 
This generates samples in a single step.
The sample quality can be improved by alternating the denoising and noise injection steps. 
In particular, in the $n$-th iteration, we first perform noise-injecting forward process to the previous predicted sample $\vz$ as 
$\hat{\vz}_{\tau_n}\sim \mathcal{N}(\alpha(\tau_{n})\vz;\sigma^2(\tau_{n})\mathbf{I})$,
where $\tau_n$ is a decreasing sequence of time steps.
This corresponds to going back to point $\hat{\vz}_{\tau_n}$ on the PF-ODE trajectory. 
Then, we perform the next $\vz_0$ prediction again using the trained latent consistency function. 
In our experiments, one can see the second iteration can already refine the generation quality significantly, and high quality images can be generated in just 2-4 steps. 
We provide the pseudo-code in Algorithm~\ref{alg:LCM_sample}. 

\begin{algorithm}[H]
    \begin{minipage}{\linewidth}
        \caption{Multistep Latent Consistency Sampling}\label{alg:LCM_sample}
        \begin{algorithmic}
            \STATE \textbf{Input:} Latent Consistency Model $\vf_\vtheta(\cdot,\cdot,\cdot,\cdot)$, Sequence of timesteps $\tau_{1}>\tau_2>\cdots >\tau_{N-1}$, Text condition $\vc$, Classifier-Free Guidance Scale $\omega$, Noise schedule $\alpha(t),\sigma(t)$, Decoder $D(\cdot)$
            \STATE Sample initial noise $\hat{\vz}_T\sim \mathcal{N}(\bm{0};\bm{I})$
            \STATE $\vz\leftarrow \vf_\vtheta(\hat{\vz}_T,\omega, \vc, T)$
            \FOR{$n=1$ to $N-1$}
            \STATE $\hat{\vz}_{\tau_n}\sim \mathcal{N}(\alpha(\tau_{n})\vz;\sigma^2(\tau_{n})\mathbf{I})$
            \STATE $\vz\leftarrow \vf_\vtheta(\hat{\vz}_{\tau_n},\omega, \vc, \tau_n)$
            \ENDFOR
        \STATE $\vx\leftarrow D(\vz)$
        \STATE \textbf{Output:} $\vx$
        \end{algorithmic} 
    \end{minipage}
\end{algorithm}

\section{Algorithm Details of Latent Consistency Fine-tuning \label{appendix:LCF}}

In this section, we provide further details of Latent Consistency Fine-tuning (LCF). The pseudo-code of LCF is provided in Algorithm~\ref{alg:LCF}. During the Latent Consistency Fine-tuning (LCF) process, we randomly select two time steps $t_n$ and $t_{n+k}$ that are $k$ time steps apart and apply the \textit{same} Gaussian noise $\boldsymbol{\epsilon}$ to obtain the noised data $\vz_{t_n},\vz_{t_{n+k}}$ as follows:
%, enforcing them in the same ODE trajectories with:
$$
\vz_{t_{n+k}} = \alpha(t_{n+k})\vz + \sigma(t_{n+k}) \boldsymbol{\epsilon}\quad, \quad \vz_{t_{n}} = \alpha(t_{n})\vz + \sigma(t_{n}) \boldsymbol{\epsilon}.
$$
Then, we can directly calculate the consistency loss for these two time steps to enforce self-consistency property in Eq.\ref{eq:self_consistency}. Notably, this method can also utilize the skipping-step technique to speedup the convergence.
%without necessitating the computation of the loss function for neighboring time steps. 
Furthermore, we note that latent consistency fine-tuning is independent of the pre-trained teacher model, facilitating direct fine-tuning of a pre-trained latent consistency model without reliance on the teacher diffusion model.

\begin{algorithm}[H]
    \begin{minipage}{\linewidth}
        \caption{Latent Consistency Fine-tuning (LCF)}\label{alg:LCF}
        \begin{algorithmic}
            \STATE \textbf{Input:} customized dataset $\mathcal{D}^{(s)}$, pre-trained LCM parameter $\vtheta$, learning rate $\eta$, distance metric $d(\cdot,\cdot)$, EMA rate $\mu$, noise schedule $\alpha(t),\sigma(t)$, {guidance scale $[w_{\text{min}},w_{\text{max}}]$, skipping interval $k$, and encoder $E(\cdot)$}
            \STATE {Encode training data into the latent space: $\mathcal{D}_z^{(s)}=\{(\vz,\vc)|\vz=E(\vx),(\vx,\vc)\in\mathcal{D}^{(s)}\}$}
            \STATE $\vtheta^-\leftarrow\vtheta$
            \REPEAT
            \STATE Sample $(\vz,\vc) \sim \mathcal{D}^{(s)}_z$, {$n\sim \mathcal{U}[1,N-k]$ and $w\sim [w_\text{min},w_\text{max}]$}
            % \STATE Sample $\vz_{t_{n+k}}\sim \mathcal{N}(\alpha(t_{n+k})\vz;\sigma^2(t_{n+k})\mathbf{I}),\vz_{t_{n}}\sim \mathcal{N}(\alpha(t_{n})\vz;\sigma^2(t_{n})\mathbf{I})$
            \STATE Sample $\boldsymbol{\epsilon} \sim \mathcal{N}\left(\mathbf{0}, \boldsymbol{I}\right)$
            \STATE $\vz_{t_{n+k}} \leftarrow \alpha(t_{n+k})\vz + \sigma(t_{n+k})\boldsymbol{\epsilon}\quad,\quad \vz_{t_{n}}\leftarrow \alpha(t_{n})\vz + \sigma(t_{n})\boldsymbol{\epsilon}$
            \STATE $\begin{aligned}\mathcal{L}(\vtheta,\vtheta^-)\leftarrow {d(\vf_\vtheta(\vz_{t_{n+k}},t_{n+k},\vc,w),\vf_{\vtheta^-}({\vz}_{t_n},t_n,\vc,w))}\end{aligned}$
          \STATE $\vtheta\leftarrow\vtheta-\eta\nabla_\vtheta\mathcal{L}(\vtheta,\vtheta^-)$
            \STATE $\vtheta^-\leftarrow \text{stopgrad}(\mu\vtheta^-+(1-\mu)\vtheta)$
            \UNTIL convergence
        \end{algorithmic} 
    \end{minipage}
\end{algorithm}

\section{Different ways to parameterize the consistency function\label{appendix:parameterization}}

As previously discussed in Eq~\ref{eq:parameterization}, we can parameterize our consistency model function $ \vf_\vtheta(\vz, \vc, t)$ in different ways, depending on the way the teacher diffusion model is parameterized. 
For $\boldsymbol{\epsilon}\text{-Prediction}$~\citep{song2020denoising}, we 
use the following parameterization:
\begin{equation}
    \vf_\vtheta(\vz, \vc, t)=c_{\text{skip}}(t)\vz+c_{\text{out}}(t)
    \hat{\vz}_0
    \quad (\boldsymbol{\epsilon}\text{-Prediction})
    \label{eq:parameterization_noise_appendix}
\end{equation}
where 
\begin{equation}
    \hat{\vz}_0 = \left(\frac{\vz_t - \sigma(t) \hat{\boldsymbol{\epsilon}}_\theta(\vz, \vc,t)}{\alpha(t)}\right).
\end{equation}
Recalling that $\vz_t = \alpha(t) \vz_0 + \sigma(t) \boldsymbol{\epsilon}$,
$\hat{\vz}_0$ can be seen as a prediction of $\vz_0$ at time $t$.

Next, we provide the parameterization of $(\boldsymbol{x}\text{-Prediction})$ \citep{ho2020denoising, salimans2022progressive} with the following form:
\begin{equation}
    \vf_\vtheta(\vz, \vc, t)=c_{\text{skip}}(t)\vz+c_{\text{out}}(t) \boldsymbol{x}_\theta(\vz_t, \vc, t),\quad (\boldsymbol{x}\text{-Prediction})
    \label{eq:parameterization_x_appendix}
\end{equation}
where $\boldsymbol{x}_\theta(\vz_t, \vc, t)$ corresponds to the teacher diffusion model with $\boldsymbol{x}$-prediction.

Finally, for $\boldsymbol{v}$-prediction \citep{salimans2022progressive}, the consistency function is parameterized as
\begin{equation}
    \vf_\vtheta(\vz, \vc, t)=c_{\text{skip}}(t)\vz+c_{\text{out}}(t)\left(\alpha_t \vz_t - \sigma_t \boldsymbol{v}_\theta(\vz_t, \vc, t)\right),\quad (\boldsymbol{v}\text{-Prediction})
    \label{eq:parameterization_v_appendix}
\end{equation}
where $\boldsymbol{v}_\theta(\vz_t, \vc, t)$ corresponds to the teacher diffusion model with $\boldsymbol{v}$-prediction.

As mentioned in Sec~\ref{sec:text-to-image}, we use the $\boldsymbol{\epsilon}\text{-Parameterization}$ in Eq.~\ref{eq:parameterization_noise_appendix} to train LCM at 512$\times$512 resolution using the teacher diffusion model, Stable-Diffusion-V2.1-Base (originally trained with $\boldsymbol{\epsilon}\text{-Prediction}$ at 512 resolution). For resolution 768$\times$768, we train the LCM using the $\boldsymbol{v}\text{-Parameterization}$ in Eq.~\ref{eq:parameterization_v_appendix}, adopting the teacher diffusion model, Stable-Diffusion-V2.1 (originally trained with $\boldsymbol{v}\text{-Prediction}$ at 768 resolution). 

\section{Formulas of Other ODE Solvers \label{appendix:solvers}}

As discussed in Sec~\ref{sec:skipping_step}, we use the DDIM \citep{song2020denoising}, DPM-Solver \citep{lu2022dpm} and DPM-Solver++ \citep{lu2022dpm++} as the PF-ODE solvers. Proven in \citep{lu2022dpm}, the DDIM-Solver is actually the first-order discretization approximation of the DPM-Solver. 

For \textbf{DDIM} \citep{song2020denoising} , the detailed formula of DDIM PF-ODE solver $\Psi_{\text{DDIM}}$ from $t_{n+k}$ to $t_n$ is provided as follows.
\begin{equation}
    \begin{aligned}
    \Psi_{\text{DDIM}}(\vz_{t_{n+k}},t_{n+k}, t_n, \vc) & = \hat{\vz}_{t_{n}} - \vz_{t_{n+k}} \\
    & = \underbrace{\frac{\alpha_{t_n}}{\alpha_{t_{n+k}}}\vz_{t_{n+k}} - \sigma_{t_n}\left(\frac{\sigma_{t_{n+k}}\cdot\alpha_{t_n}}{\alpha_{t_{n+k}}\cdot\sigma_{t_n}} - 1\right)\hat{\boldsymbol{\epsilon}}_\theta(\vz_{t_{n+k}},\vc,t_{n+k})}_{\text{DDIM Estimated} \ \vz_{t_{n}}} - \vz_{t_{n+k}}
    \end{aligned}
\end{equation}

For \textbf{DPM-Solver} \citep{lu2022dpm}, we only consider the case for $order=2$, and the detailed formula of PF-ODE solver $\Psi_{\text{DPM-Solver}}$ is provided as follows. 
First we define some notations. 
We denote $\lambda_{t_n}=\log(\frac{\alpha_{t_n}}{\sigma_{t_n}})$, which is the Log-SNR, $h_{t_n}^0=\lambda_{t_n}-\lambda_{t_{n+k}}, h_{t_n}^1=\lambda_{t_n}-\lambda_{t_{n+k/2}}$, and $r_{t_n}= h_{t_n}^1 / h_{t_n}^0$.
\begin{equation}
    \begin{aligned}
    & \Psi_{\text{DPM-Solver}}(\vz_{t_{n+k}},t_{n+k}, t_n, \vc)  \\
    & = \frac{\alpha_{t_n}}{\alpha_{t_{n+k}}}\vz_{t_{n+k}} - \sigma_{t_n}(e^{h_{t_n}^0} - 1)\hat{\boldsymbol{\epsilon}}_\theta(\vz_{t_{n+k}},\vc,t_{n+k}) \\
    & - \frac{\sigma_{t_n}}{2r_{t_n}}(e^{h_{t_n}^0}-1)\left(\hat{\boldsymbol{\epsilon}}_\theta(\vz^\Psi_{t_{n+k/2}},\vc,t_{n+k/2})-\hat{\boldsymbol{\epsilon}}_\theta(\vz_{t_{n+k}},\vc,t_{n+k})\right) - \vz_{t_{n+k}},
    \end{aligned}
\end{equation}
where $\hat{\boldsymbol{\epsilon}}$ is the noise prediction model, and $\vz^\Psi_{t_{n+k/2}}$ is the middle point between $n+k$ and $n$, given by the following formula:
\begin{equation}
    \begin{aligned}
        \vz^\Psi_{t_{n+k/2}} &= \frac{\alpha_{t_{n+k/2}}}{\alpha_{t_{n+k}}}\vz_{t_{n+k}} - \sigma_{t_{n+k/2}}(e^{h_{t_n}^1} - 1)\hat{\boldsymbol{\epsilon}}_\theta(\vz_{t_{n+k}},\vc,t_{n+k})
    \end{aligned}
\end{equation}

For \textbf{DPM-Solver++} \citep{lu2022dpm++}, we consider the case for $order=2$, DPM-Solver++ replaces the original noise prediction to data prediction \citep{lu2022dpm++}, with the detailed formula of $\Psi_{\text{DPM-Solver++}}$ provided as follows.
\begin{equation}
    \begin{aligned}
    & \Psi_{\text{DPM-Solver++}}(\vz_{t_{n+k}},t_{n+k}, t_n, \vc)  \\
    & = \frac{\sigma_{t_n}}{\sigma_{t_{n+k}}}\vz_{t_{n+k}} - \alpha_{t_n}(e^{-h_{t_n}^0} - 1)\hat{\boldsymbol{x}}_\theta(\vz_{t_{n+k}},\vc,t_{n+k}) \\
    & - \frac{\alpha_{t_n}}{2r_{t_n}}(e^{-h_{t_n}^0}-1)\left(\hat{\boldsymbol{x}}_\theta(\vz^\Psi_{t_{n+k/2}},\vc,t_{n+k/2})-\hat{\boldsymbol{x}}_\theta(\vz_{t_{n+k}},\vc,t_{n+k})\right) - \vz_{t_{n+k}},
    \end{aligned}
\end{equation}
where $\hat{\boldsymbol{x}}$ is the data prediction model \citep{lu2022dpm} and $\vz^\Psi_{t_{n+k/2}}$ is the middle point between $n+k$ and $n$, given by the following formula:
\begin{equation}
    \begin{aligned}
        \vz^\Psi_{t_{n+k/2}} &= \frac{\sigma_{t_{n+k/2}}}{\sigma_{t_{n+k}}}\vz_{t_{n+k}} - \alpha_{t_{n+k/2}}(e^{-h_{t_n}^1} - 1)\hat{\boldsymbol{x}}_\theta(\vz_{t_{n+k}},\vc,t_{n+k})
    \end{aligned}
\end{equation}

\section{Training Details of Latent Consistency Distillation \label{appendix:training_details}}

As mentioned in Section~\ref{sec:text-to-image}, we conduct our experiments in two resolution settings 512$\times$512 and 768$\times$768. For the former setting, we use the LAION-Aesthetics-6+ \citep{schuhmann2022laion} 12M dataset, consisting of 12M text-image pairs with predicted aesthetics scores higher than 6. For the latter setting, we use the LAIOIN-Aesthetic-6.5+ \citep{schuhmann2022laion}, which comprise 650K text-image pairs with predicted aesthetics scores higher than 6.5.

For 512$\times$512 resolution, we train the LCM with the teacher diffusion model Stable-Diffusion-V2.1-Base (SD-V2.1-Base) \citep{rombach2022high}, which is originally trained on 512$\times$512 resolution images using the $\boldsymbol{\epsilon}$-Prediction \citep{ho2020denoising}. We train LCM (512$\times$512) with 100K iterations on 8 A100 GPUs, using a batch size of 72, the same learning rate 8e-6 , EMA rate $\mu=0.999943$ and Rectified Adam optimizer \citep{liu2019variance} used in \citep{song2023consistency}. We select the DDIM-Solver \citep{song2020denoising} and skipping step $k=20$ in Eq.~\ref{eq:aug_pf_ode_final}. We set the guidance scale range $[\omega_{\text{min}}, \omega_{\text{max}}] = [2, 14]$, which is consistent with the setting in Guided-Distill \citep{meng2023distillation}. During training, we initialize the consistency function $\boldsymbol{f_\theta}(\boldsymbol{\vz}_{t_{n}}, \omega, \boldsymbol{c}, t_{n})$ with the same parameters as the teacher diffusion model (SD-V2.1-Base). To encode the CFG scale $\omega$ into the LCM, we applying Fourier embedding to $\omega$, integrating it into the origin LCM backbone by adding the projected $\omega$-embedding into the original embedding, as done in \citep{meng2023distillation}.  We use a zero parameter initialization method mentioned in \citep{zhang2023adding} on projected $\omega$-embedding for better training stability. For training LCM (512$\times$512), we use a augmented consistency function parameterized in $\boldsymbol{\epsilon}$-prediction as discussed in Appendix.~\ref{appendix:parameterization}.

For 768$\times$768 resolution, we train the LCM with the teacher diffusion model Stable-Diffusion-V2.1 (SD-V2.1) \citep{rombach2022high}, which is originally trained on 768$\times$768 resolution images using the $\boldsymbol{v}$-Prediction \citep{salimans2022progressive}. We train LCM (768$\times$768) with 100K iterations on 8 A100 GPUs using a batch size of 16, while the other hyper-parameters remain the same as in 512$\times$512 resolution setting.

\section{Reproduction Details of Guided-Distill \label{appendix:reproduction}}

Guided-Distill \citep{meng2023distillation} serves as a significant baseline for guided distillation but is not open-sourced. We adhered strictly to the training procedure described in the paper, reproducing the method for accurate comparisons. For 512$\times$512 resolution setting, Guided-Distill \citep{meng2023distillation} used a large batch size of 512, which requires at least 32 A100 GPUs for training. Due to limited resource, we reduced the batch size to 72 (512 resolution), while set the batchsize to 16 for 768 resolution, the same as ours, and trained for 100K iterations, also the same as in LCM. 

Specifically, Guided Distill involves two stages of distillation. For the first stage, it use a student model to fit the outputs of the pre-trained guided diffusion model using classifier-free guidance scales $\omega$. The loss function is as follows:
\begin{equation}
    \mathbb{E}_{w\sim p_w,t\sim \mathcal{U}[0,1],\vx\sim p_{\text{data}}}(\vx)[\omega(\lambda_t)||\hat{\vx}_{\bm{\eta}_1}(\vz_t,w)-\hat{\vx}_\vtheta^w(\vz_t)||_2^2],
\end{equation}
where $\hat{\vx}_\vtheta(\vz_t)=(1+w)\hat{\vx}_{c,\vtheta}(\vz_t)-w\hat{\vx}_\vtheta(\vz_t),\vz_t\sim q(\vz_t|\vx)$ and $p_w(w)=\mathcal{U}[w_{\text{min}},w_{\text{max}}]$. 

In our implementation, we follow the same training procedure in \citep{meng2023distillation} except the difference of computation resources. For \textbf{first stage distillation}, we train the student model with 25,000 gradient updates (batch size 72), roughly the same computation costs in \citep{meng2023distillation} (3,000 gradient updates, batch size 512), and we reduce the original learning rate $1e-4$ to $5e-5$ for smaller batch size. For \textbf{second stage distillation}, we progressively train the student model using the same schedule as in Guided-Distill \citep{meng2023distillation} except for batch size difference.  We train the student model with 2500 gradient updates except when the sampling step equals to 1,2, or 4, where we train for 20000 gradient updates, the same schedule used in \citep{meng2023distillation}. We trained until the total number of gradient iterations for the entire stage reached 100K the same as LCM training. The generation results of Guided Distill are shown in Figure~\ref{fig:text_to_image}.
We can also see that the performances in Table~\ref{tab:512_results} and Table~\ref{tab:768_results} are similar, further verifying the correctness of our Guided-Distill implementation. Nevertheless, we acknowledge that longer training and more computational resources can lead to better results as reported in \citep{meng2023distillation}. However, LCM achieves faster convergence and superior results under the same computation cost (same batch size, same number of iterations), demonstrating its practicability and superiority.

\section{More few-step Inference Results \label{appendix:more_results}}

\begin{figure}[t] 
\begin{centering}
\includegraphics[scale=0.82]{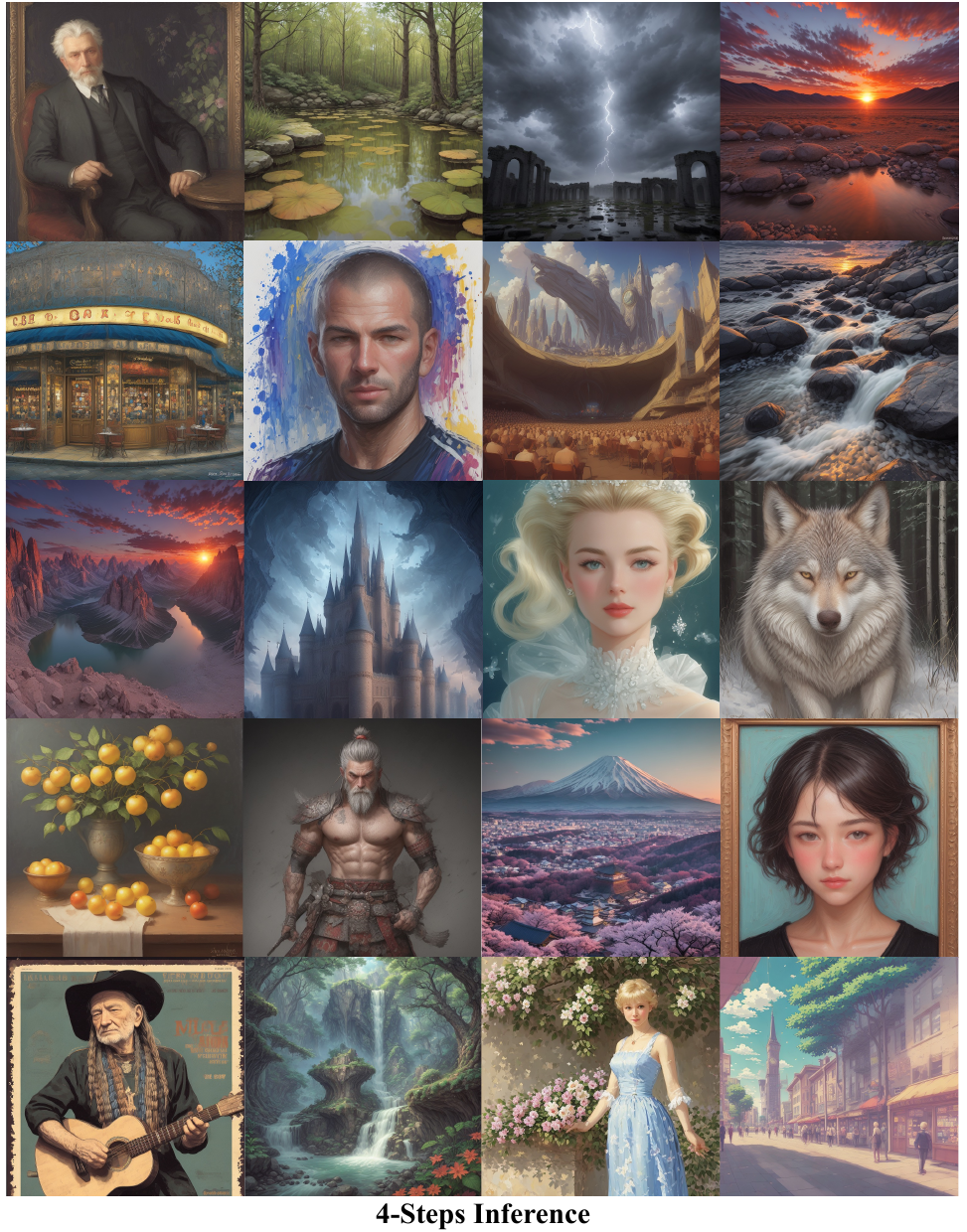}
 \vspace{-0.1in}
\caption{More generated images results with LCM 4-steps inference (768$\times$768 Resolution). We employ LCM to distill the Dreamer-V7 version of SD in just 4,000 training iterations.}\label{fig:more_results_4steps}
\end{centering}
\vspace{-0.1in}
\end{figure}

\begin{figure}[t] 
\begin{centering}
\includegraphics[scale=0.82]{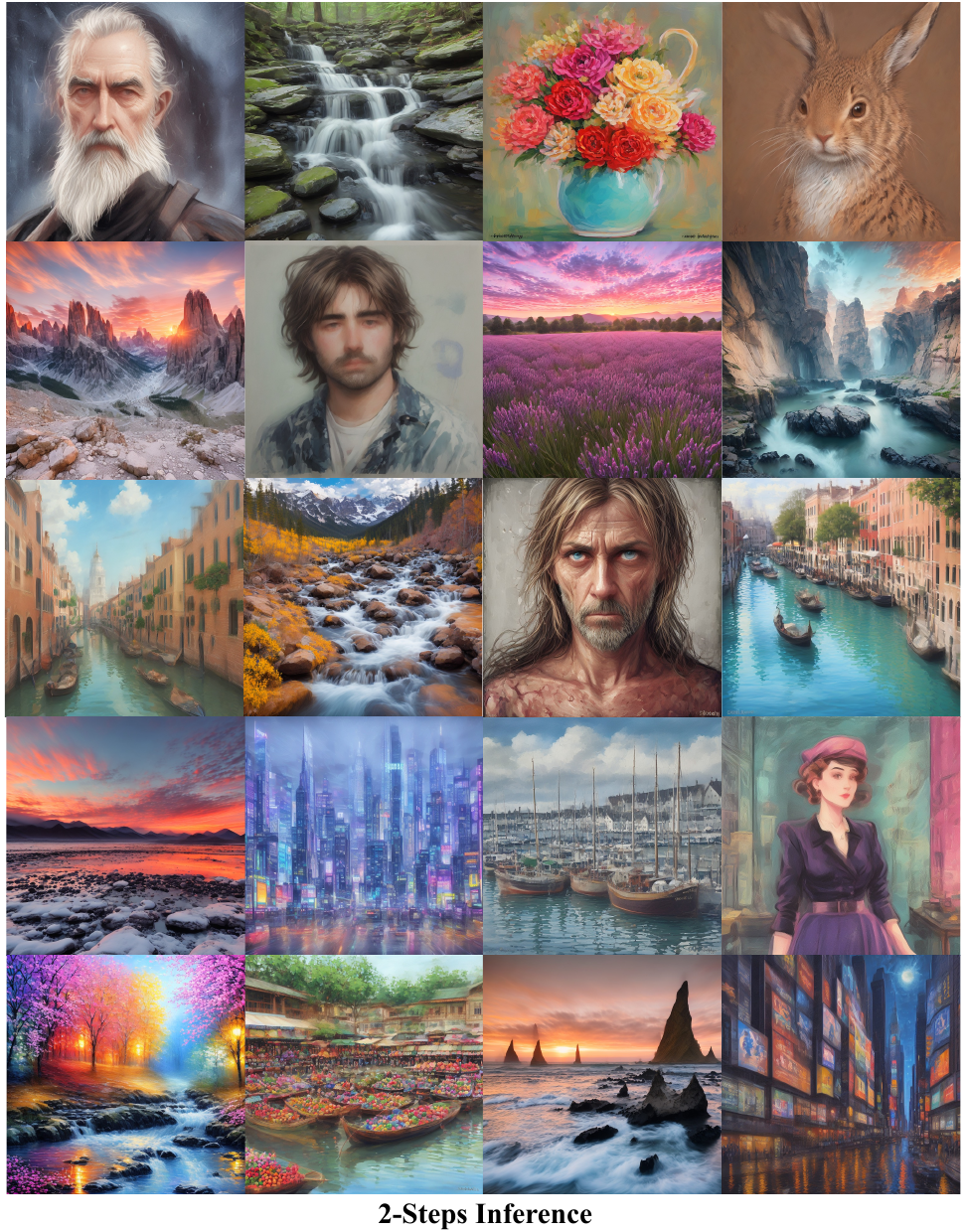}
 \vspace{-0.1in}
\caption{More generated images results with LCM 2-steps inference (768$\times$768 Resolution). We employ LCM to distill the Dreamer-V7 version of SD in just 4,000 training iterations.}\label{fig:more_results_2steps}
\end{centering}
\vspace{-0.1in}
\end{figure}

We present more images (768$\times$768) generation results with LCM using 4 and 2-steps inference in Figure~\ref{fig:more_results_4steps} and Figure~\ref{fig:more_results_2steps}.  It is evident that LCM is capable of synthesizing high-resolution images with just 2, 4 steps of inference. Moreover, LCM can be derived from any pre-trained Stable Diffusion (SD) \citep{rombach2022high} in merely 4,000 training steps, equivalent to around 32 A100 GPU Hours, showcasing the effectiveness and superiority of LCM.

\end{document}